# A Kriging-HDMR-based surrogate model with sample pool-free active learning strategy for reliability analysis

Wenxiong Li*, Hanyu Liao, Suiyin Chen

College of Water Conservancy and Civil Engineering, South China Agricultural University, Guangzhou 510642, China

E-mail: leewenxiong@scau.edu.cn

**ABSTRACT**

In reliability engineering, conventional surrogate models encounter the "curse of dimensionality" as the number of random variables increases. While the active learning Kriging surrogate approaches with high-dimensional model representation (HDMR) enable effective approximation of high-dimensional functions and are widely applied to optimization problems, there are rare studies specifically focused on reliability analysis, which prioritizes prediction accuracy in critical regions over uniform accuracy across the entire domain. This study develops an active learning surrogate model method based on the Kriging-HDMR modeling for reliability analysis. The proposed approach facilitates the approximation of high-dimensional limit state functions through a composite representation constructed from multiple low-dimensional sub-surrogate models. The architecture of the surrogate modeling framework comprises three distinct stages: developing single-variable sub-surrogate models for all random variables, identifying the requirements for coupling-variable sub-surrogate models, and constructing the coupling-variable sub-surrogate models. Optimization mathematical models for selection of design of experiment samples are formulated based on each stage's characteristics, with objectives incorporating uncertainty variance, predicted mean, sample location and inter-sample distances. A candidate sample pool-free approach is adopted to achieve the selection of informative samples. Numerical experiments demonstrate that the proposed method achieves high computational efficiency while maintaining strong predictive accuracy in solving high-dimensional reliability problems.

## 1 Introduction

Engineering systems are inherently exposed to uncertainties (e.g., material variability, loading conditions, geometric deviations), which can critically degrade reliability and safety. Consequently, reliability analysis is indispensable across disciplines (mechanical, civil, automotive, robotics). Traditional approaches fall into two paradigms: gradient-based methods (e.g., First/Second Order Reliability Methods, FORM/SORM) and sampling-based techniques (e.g., Monte Carlo Simulation, MCS, Importance Sampling, IS).

FORM/SORM determine the Most Probable Point (MPP) and reliability index by performing low-order Taylor expansions of the Limit State Function (LSF), offering computational efficiency but requiring strict LSF regularity and neglecting high-order effects. As a consequence, they struggle when dealing with nonlinear LSFs or system-level problems. Sampling methods relax regularity constraints but demand extensive random samples (e.g., MCS incurs high time costs due to per-sample LSF evaluations; IS improves efficiency by targeting critical regions). For highly nonlinear or small-failure-probability scenarios, surrogate model-based methods have emerged as a solution by replacing the expensive LSF with a data-driven approximation (e.g., Kriging [1], support vector machine [2], radial basis function [3, 4] and neural networks [5]), significantly enhancing computational tractability.

Among surrogates, Kriging dominates reliability analysis due to its interpolation accuracy and uncertainty quantification (via prediction variance), which guides sample selection [6, 7]. The core challenge lies in constructing the surrogate using Design of Experiments (DoE) data. Poor-quality DoE, which often results from traditional sampling, leads to inefficient models. To address this issue, Active Learning (AL) strategies have been developed. These strategies dynamically select informative samples by iteratively evaluating uncertainty using learning functions, and optimizes the surrogate iteratively. Correspondingly, various learning functions have been proposed for the AL strategies, such as the Expected Feasibility Function (EFF) [6], the sign indication learning function (U-learning function) [1], the information entropy theory-based learning function (H-function) [8], the Least Improvement Function (LIF) [9], the reliability-based active learning function (REIF) [15], the cross-validation-based learning function (C-function) [10], the active weight learning function [11], the distance-based function [12] and the Sample-based Expected Uncertainty Reduction (SEUR) learning function [13]. Based on the AL strategies, there are also extensions for small failures [7, 14, 15] and for multiple failure modes [16-20]. In most AL-based surrogated model methods, the DoE samples are obtained from a Candidate Sample Pool (CSP). However, the size of CSP critically impacts performance: larger CSPs enhance solution accuracy by increasing the likelihood of including high informative samples, but they also significantly escalate computational costs due to the sheer number of evaluations required. Conversely, smaller CSPs reduce computational burden but may fail to capture essential samples, particularly those critical for identifying rare failure events or complex system behaviors. As a result, CSP-free AL surrogate model methods have been proposed. These methods, which use optimization techniques like Particle Swarm Optimization (PSO) and Genetic Algorithm (GA) to directly identify informative samples without predefined pools, include the work of Li et al. [21], Jing et al. [4], and Meng et al. [11].

In practical engineering, systems often involve a large number of random variables, resulting in high-dimensional LSFs. The curse of dimensionality exacerbates this issue: the required number of DoE samples grows exponentially with dimensionality, making traditional surrogate model methods impractical. To tackle this, dimensionality reduction techniques, such as High-Dimensional Model Representation (HDMR) [22, 23], have been developed. HDMR decomposes a high-dimensional function into a sum of low-dimensional component functions, leveraging reference points to capture interactions efficiently. Prior studies on high-dimensional function approximation and optimization have demonstrated HDMR's potential in reducing computational costs. Li et al. [24] proposed a projection-based intelligent sampling method with the HDMR for approximation of nonlinear functions. Meng et al. [25] proposed a modeling method that combines the Kriging model with HDMR to achieve efficient approximation of nonlinear response functions. Cai et al. [26] developed a new model by combining HDMR with an enhanced radial basis function, and the proposed model was used in structural design optimization. Cai et al. [27] proposed a hybrid metamodel which combines HDMR with Co-Kriging and Kriging, and the proposed method can efficiently use multi-fidelity samples to approximate black-box problems through a two-stage metamodeling strategy. Li et al. [28] developed the SVR-based HDMR to efficiently and effectively construct the HDMR expansion by Support Vector Regression (SVR). Li et al. [29] improved the efficiency of the Kriging-HDMR model by introducing Pareto frontier expected improvement. Wu et al. [30] combined the Teaching-Learning-Based Optimization (TLBO) algorithm with HDMR, where the TLBO algorithm was utilized to search for high-quality DoE for model construction, and the proposed method was applied to high-dimensional aerodynamic shape optimization. Yue et al. [31] combined Polynomial Chaos Expansion (PCE) with HDMR to construct the PCE-HDMR surrogate model. Kim and Lee [32] proposed the selectively high-order Kriging-HDMR for efficient metamodeling of high-dimensional, expensive, blackbox problems, where high-order interaction components were selected for surrogate model construction. Zhang et al. [33] developed a PC-Kriging-HDMR approximate modeling method by embedding the PC-Kriging surrogate model into the framework of HDMR.

Existing HDMR-based surrogate studies primarily focus on high-dimensional function approximation and optimization problems, with scarce reliability-focused research. Although HDMR-based surrogate models can theoretically replace structural responses for both optimization and failure probability calculations, optimization demands uniform high accuracy across the entire variable domain to avoid local optima, whereas reliability analysis focuses on precise limit-state prediction (with other regions requiring only correct sign classification for failure/safety distinction). In other words, reliability analysis prioritizes prediction accuracy in critical regions over uniform accuracy across the entire domain. This disparity in objectives and characteristics directly necessitates distinct active learning strategies for surrogate models in the two application scenarios. To the best of the authors' knowledge, apart from a few scattered studies such as [34], research specifically integrating HDMR with AL surrogate model methods for reliability analysis remains scarce. The performance of such methods remains difficult to evaluate accurately, highlighting the need for dedicated investigation, which forms the research objective of this study. For high-dimensional active learning, traditional CSP-based strategies require large candidate pools to ensure representativeness in multi-variable sampling spaces, which risks omission of key samples. In contrast, CSP-free approachs directly target optimal samples without depending on large candidate sets, avoiding scalability issues and sample loss risks. Therefore, for high-dimensional reliability analysis problems, HDMR-based AL surrogate model with a CSP-free approach can be expected to have high performance.

This study develops an AL surrogate model method based on Kriging-HDMR modeling for reliability analysis. The proposed approach facilitates the approximation of high-dimensional LSFs through a composite representation constructed from multiple low-dimensional sub-surrogate models. The architecture of the surrogate modeling framework comprises three distinct stages: developing single-variable sub-surrogate models for all random variables, identifying the requirements for coupling-variable sub-surrogate models, and constructing the coupling-variable sub-surrogate models. Regarding the selection of DoE samples, corresponding optimization mathematical models are formulated based on the characteristics of each modeling stage, with optimization objectives incorporating the effects of uncertainty variance, predicted mean, sample point location and inter-sample distances. In addition, the CSP-free approach is adopted to achieve the selection of informative samples. Finally, numerical experiments validate the performance of the proposed surrogate model method in addressing high-dimensional reliability problems.

## 2 Kriging-HDMR model

2.1 Kriging model

The Kriging interpolation framework, originally developed within the domain of geostatistical analysis for spatial data estimation, has subsequently evolved into a prominent computational metamodel in reliability engineering. As a surrogate model, it effectively approximates the functional relationship between input variables and output responses, particularly in characterizing the correlation between sampling points and their LSF evaluations.

Kriging model fundamentally incorporates two components: (1) a deterministic parametric regression component that establishes the mean response function through linear parameter estimation, and (2) a nonparametric stochastic process component manifesting as a correlated Gaussian random field. Then, the approximate relationship between any experiment $\mathbf{x} = \{x_1, x_2, \ldots, x_n\}^T$ and the response $\hat{G}(\mathbf{x})$ can be expressed as

$$\hat{G}(\mathbf{x}) = \mathbf{f}^T(\mathbf{x})\boldsymbol{\beta}_r + z(\mathbf{x}) \tag{1}$$

where $\mathbf{f}^T(\mathbf{x})\boldsymbol{\beta}$ represents the deterministic part of the mean response approximation, consisting of $k$ basis functions in the vector $\mathbf{f}(\mathbf{x}) = \{f_1(\mathbf{x}), f_2(\mathbf{x}), \ldots, f_k(\mathbf{x})\}^T$ and the corresponding regression coefficients in the vector $\boldsymbol{\beta}_r = \{\beta_{r1}, \beta_{r2}, \ldots, \beta_{rk}\}^T$. In this paper, ordinary Kriging model is selected, namely $f_i(\mathbf{x}) = 1 \ (i = 1, 2, \ldots, k)$. In Eq. (1), $z(\mathbf{x})$ is a stationary Gaussian process with zero mean $z(\mathbf{x}) \sim N(0, \sigma^2)$, and the covariance between two points of space $\mathbf{x}^i$ and $\mathbf{x}^j$ is defined as

$$\mathrm{Cov}\left[z(\mathbf{x}^i), z(\mathbf{x}^j)\right] = \sigma^2 R(\boldsymbol{\theta}, \mathbf{x}^i, \mathbf{x}^j) \tag{2}$$

where $\sigma^2$ is the process variance, $\boldsymbol{\theta} = \{\theta_1, \theta_2, \ldots, \theta_{N_D}\}$ refers to the parameter vector where $N_D$ corresponds to the number of random variables, and $R(\boldsymbol{\theta}, \mathbf{x}^i, \mathbf{x}^j)$ represents the correlation function between $\mathbf{x}^i$ and $\mathbf{x}^j$, which is formulated by

$$R(\boldsymbol{\theta}, \mathbf{x}^i, \mathbf{x}^j) = \prod_{d=1}^{N_D} \exp\left[-\theta_d \left(x_d^i - x_d^j\right)\right] \tag{3}$$

where $x_d^i$ and $x_d^j$ refer to the $d$-th components in $\mathbf{x}^i$ and $\mathbf{x}^j$, respectively.

For a given set of DoE $\mathbf{S}_{\mathrm{DoE}} = [\mathbf{x}^1, \mathbf{x}^2, \ldots, \mathbf{x}^m]$ with $m$ being the number of samples in DoE, the corresponding response set is denoted as $\mathbf{Y}_{\mathrm{DoE}} = [G(\mathbf{x}^1), G(\mathbf{x}^2), \ldots, G(\mathbf{x}^m)]$. Then, the scalars $\beta_r$ and $\sigma^2$ are estimated by

$$\hat{\beta}_r = \left(\mathbf{1}^T \mathbf{R}^{-1} \mathbf{1}\right)^{-1} \mathbf{1}^T \mathbf{R}^{-1} \mathbf{Y}_{\mathrm{DoE}} \tag{4}$$

$$\hat{\sigma}^2 = \frac{1}{m} \left(\mathbf{Y}_{\mathrm{DoE}} - \hat{\beta}_r \mathbf{1}\right)^T \mathbf{R}^{-1} \left(\mathbf{Y}_{\mathrm{DoE}} - \hat{\beta}_r \mathbf{1}\right) \tag{5}$$

where $\mathbf{R}$ refers to the correlation matrix with the component $R_{i,j} = R(\boldsymbol{\theta}, \mathbf{x}^i, \mathbf{x}^j)$ representing the correlation between each pair of points in DoE, and $\mathbf{1}$ refers to the vector filled with 1 of length $m$. $\hat{\beta}_r$ and $\hat{\sigma}^2$ in Eqs. (4) and (5) are related to the correlation parameters $\theta_i$ through the matrix $\mathbf{R}$, then the value of $\boldsymbol{\theta}$ is required to obtained by using maximum likelihood estimation:

$$\boldsymbol{\theta} = \arg\min_{\theta} \left(\det(\mathbf{R})\right)^{\frac{1}{m}} \hat{\sigma}^2 \tag{6}$$

According to the Gaussian process regression theory, the system response follows the normal distribution as $G(\mathbf{x}) \sim N(\mu_G(\mathbf{x}), \sigma_G(\mathbf{x}))$. Then, based on the Kriging model established according to the given data of DoE, the best linear unbiased predictor of the response $\hat{G}(\mathbf{x})$ at an unknown $\mathbf{x}$ is shown to be a Gaussian random variate $\hat{G}(\mathbf{x}) \sim N(\mu_{\hat{G}}(\mathbf{x}), \sigma_{\hat{G}}(\mathbf{x}))$ where $\mu_{\hat{G}}(\mathbf{x})$ and $\sigma_{\hat{G}}(\mathbf{x})$ are

$$\mu_{\hat{G}}(\mathbf{x}) = \hat{\beta}_r + \mathbf{r}(\mathbf{x})\mathbf{R}^{-1}\left(\mathbf{Y}_{\text{DoE}} - \hat{\beta}_r \mathbf{1}\right) \tag{7}$$

$$\sigma_{\hat{G}}^2(\mathbf{x}) = \hat{\sigma}^2\left(1 - \mathbf{r}^{\text{T}}(\mathbf{x})\mathbf{R}^{-1}\mathbf{r}(\mathbf{x}) + u^{\text{T}}(\mathbf{x})\left(\mathbf{1}^{\text{T}}\mathbf{R}^{-1}\mathbf{1}\right)^{-1}u(\mathbf{x})\right) \tag{8}$$

where $\mathbf{r}(\mathbf{x}) = \left[R(\boldsymbol{\theta},\mathbf{x},\mathbf{x}^1), R(\boldsymbol{\theta},\mathbf{x},\mathbf{x}^2), \ldots, R(\boldsymbol{\theta},\mathbf{x},\mathbf{x}^m)\right]^{\text{T}}$ and $u(\mathbf{x}) = \mathbf{1}^{\text{T}}\mathbf{R}^{-1}\mathbf{r}(\mathbf{x}) - 1$.

In Kriging model, the predicted mean value at any point $\mathbf{x}^{(i)}$ in DoE is consistent with the real response value, namely $\mu_{\hat{G}}(\mathbf{x}^i) = G(\mathbf{x}^i)$ $(i = 1, 2, \ldots, m)$, and the corresponding Kriging variance is null, namely $\sigma_{\hat{G}}^2(\mathbf{x}^i) = 0$ $(i = 1, 2, \ldots, m)$. For any point out of DoE, the Kriging variance is not zero, and its value reflects the accuracy of the prediction results at the point.

2.2 High-dimensional model representation

Two primary categories of HDMR are the Analysis Of Variance (ANOVA)-HDMR and Cut-HDMR [22, 23]. ANOVA-HDMR, originally developed for statistical applications, excels in identifying key variables and correlations, making it particularly advantageous for sensitivity analysis. However, a notable drawback of ANOVA-HDMR is its computational demand, as it requires the evaluation of numerous integrals, which are typically approximated through MCS. In contrast, Cut-HDMR eliminates the need for integral computations by approximating the target function through a decomposition into component functions defined on lines, planes, and hyperplanes that intersect at a chosen reference point, commonly referred to as the cut point. Due to its straightforward implementation and computational efficiency, Cut-HDMR is often considered more appealing for model development and application. Therefore, Cut-HDMR is employed in this work, and it is directly denoted as HDMR in this paper for the sake of convenience.

In HDMR, $f(\mathbf{x})$ can be represented as a hierarchical correlated function expansion with respect to an $n$-dimensional input vector $\mathbf{x} = [x_1, x_2, \cdots, x_n]^{\text{T}} \in \mathbb{R}^n$ as

$$f(\mathbf{x}) = f_0 + \sum_{1 \leq i \leq n} f_i(x_i) + \sum_{1 \leq i < j \leq n} f_{ij}(x_i, x_j) + \sum_{1 \leq i < j < k \leq n} f_{ijk}(x_i, x_j, x_k) + \cdots \\ + \sum_{1 \leq i < j < \ldots r \leq n} f_{ij\ldots r}(x_i, x_j, \ldots, x_r) + \cdots + f_{ij\ldots n}(x_i, x_j, \ldots, x_n) \tag{9}$$

where $f_0$ represents a constant term that corresponds to the zeroth-order contribution to $f(\mathbf{x})$, $f_i(x_i)$ denotes a first-order term that captures the independent influence of the variable $x_i$, $f_{ij}(x_i, x_j)$ signifies a second-order term that accounts for the correlated effect between variables $x_i$ and $x_j$, after removing their individual contributions, the higher-order terms describe the effects of increasing numbers of interacting variables, and the final term, $f_{ij\ldots n}(x_i, x_j, \ldots, x_n)$, represents the residual effect of all input variables, after removing all lower-order correlations and individual influences. Consequently, for a cut point $\mathbf{x}_0 = \left[x_{1_0}, x_{2_0}, \ldots, x_{n_0}\right]^{\text{T}}$ within the input variable space, the component functions of HDMR, as shown in Eq.(9), can be represented as

$$\begin{aligned}
&f_0 = f(\mathbf{x}_0) \\
&f_i(x_i) = f(x_i, \mathbf{x}_0^i) - f_0 \\
&f_{ij}(x_i, x_j) = f(x_i, x_j, \mathbf{x}_0^{ij}) - f_i(x_i) - f_j(x_j) - f_0 \\
&f_{ijk}(x_i, x_j, x_k) = f(x_i, x_j, x_k, \mathbf{x}_0^{ijk}) - f_{ij}(x_i, x_j) - f_{jk}(x_j, x_k) - f_{ik}(x_i, x_k) - f_i(x_i) - f_j(x_j) - f_k(x_k) - f_0 \\
&\ldots \\
&f_{ij\ldots n}(x_i, x_j, \ldots, x_n) = f(\mathbf{x}) - f_0 - \sum_{1 \leq i \leq n} f_i(x_i) - \sum_{1 \leq i < j \leq n} f_{ij}(x_i, x_j) - \cdots
\end{aligned} \quad (10)$$

where $\mathbf{x}_0^i$, $\mathbf{x}_0^{ij}$ and $\mathbf{x}_0^{ijk}$ are $\mathbf{x}_0$ without elements $x_i$, $(x_i, x_j)$ and $(x_i, x_j, x_k)$, respectively. The points $\mathbf{x}_0$, $(x_i, \mathbf{x}_0^i) = [x_{1_0}, x_{2_0}, \ldots, x_i, \ldots, x_{n_0}]^T$ and $(x_i, x_j, \mathbf{x}_0^{ij}) = [x_{1_0}, x_{2_0}, \ldots, x_i, \ldots, x_j, \ldots, x_{n_0}]^T$ are referred as the zeroth-order, first-order and second-order points, respectively. Accordingly, $f(\mathbf{x}_0)$, $f(x_i, \mathbf{x}_0^i)$ and $f(x_i, x_j, \mathbf{x}_0^{ij})$ are the values of $f(\mathbf{x})$ at points $\mathbf{x}_0$, $(x_i, \mathbf{x}_0^i)$ and $(x_i, x_j, \mathbf{x}_0^{ij})$, respectively, while $f_i(x_i)$ is the first-order component output response at $x_i$ along the $i$-th cut line and $f_{ij}(x_i, x_j)$ is the second-order component output response at $(x_i, x_j)$ on the $i$-$j$-th cut plane.

For the majority of physical systems, the higher-order terms in such expansions are often negligible, as only low-order correlations among input variables exert significant influence on the output response, as noted by Rabitz and Alis [23]. Consequently, the second-order HDMR expansion is widely adopted due to its ability to substantially reduce computational demands while maintaining acceptable modeling accuracy, particularly when constructing input-output mappings for complex physical systems.

2.3 Kriging surrogate with HDMR

In reliability analysis, the random variables represented in the original design space can be derived from those in the standard normal space via the Nataf transformation $\mathbf{X} = T(\mathbf{U})$. The implementation of this transformation can be accomplished by referring to the open-source package FERUM [35]. The correlation between the standard normal space and the original design space is illustrated in **Fig. 1**. Based on the standard normal space, the significance of the distance between two sample points is not contingent upon the actual distribution of random variables. This makes it more suitable for defining the distance parameters in selection of DoE samples. In the standard normal space, a point on the Limit State Equation (LSE, $G(\mathbf{u}) = 0$) that is closer to the origin represents more crucial information for estimating the real failure probability, such as the MPP shown in **Fig. 1**.

With the introduction of HDMR, the surrogate model $\hat{G}(\mathbf{u})$ for the LSF $G(\mathbf{u})$, which is a high-dimensional function of $\mathbf{u} = (u_1, u_2, \ldots, u_{N_D})(N_D \geq 3)$, can be expressed by the following second-order/third-order combinations as

$$\hat{G}(\mathbf{u}) = G(\mathbf{u}_0) + \sum_{1 \leq i \leq N_D} \bar{G}_i(u_i) + \sum_{1 \leq i < j \leq N_D} \bar{G}_{ij}(u_i, u_j) \quad \text{(Second-order combination)} \quad (11)$$

$$\hat{G}(\mathbf{u}) = G(\mathbf{u}_0) + \sum_{1 \leq i \leq N_D} \bar{G}_i(u_i) + \sum_{1 \leq i < j \leq N_D} \bar{G}_{ij}(u_i, u_j) + \sum_{1 \leq i < j < k \leq N_D} \bar{G}_{ijk}(u_i, u_j, u_k) \quad \text{(Third-order combination)} \quad (12)$$

where $\mathbf{u}_0 = (u_{1_0}, u_{2_0}, \ldots, u_{N_{D0}})$ represents the cut point within the input variable space, and

$$\bar{G}_i(u_i) = \hat{G}_i(u_i) - G(\mathbf{u}_0) \tag{13}$$

$$\bar{G}_{ij}(u_i, u_j) = \hat{G}_{ij}(u_i, u_j) - \bar{G}_i(u_i) - \bar{G}_j(u_j) - G(\mathbf{u}_0) \tag{14}$$

$$\bar{G}_{ijk}(u_i, u_j, u_k) = \hat{G}_{ijk}(u_i, u_j, u_k) - \bar{G}_{ij}(u_i, u_j) - \bar{G}_{jk}(u_j, u_k) - \bar{G}_{ik}(u_i, u_k) - \bar{G}_i(u_i) - \bar{G}_j(u_j) - \bar{G}_k(u_k) - G(\mathbf{u}_0) \tag{15}$$

In Eqs. (13)-(15), $\hat{G}_i(u_i)$, $\hat{G}_{ij}(u_i, u_j)$ and $\hat{G}_{ijk}(u_i, u_j, u_k)$ represent the low-dimensional sub-surrogate models with one, two and three input variables, for the functions $G(u_i, \mathbf{u}_0^i)$, $G(u_i, u_j, \mathbf{u}_0^{ij})$ and $G(u_i, u_j, u_k, \mathbf{u}_0^{ijk})$, respectively. Therefore, it is sufficient to construct low-dimensional sub-surrogate models such as $\hat{G}_i(u_i)$, $\hat{G}_{ij}(u_i, u_j)$ and $\hat{G}_{ijk}(u_i, u_j, u_k)$ to achieve an approximate surrogate for the high-dimensional function $G(\mathbf{u})$. The sample datasets including the input and output data for establishing the low-dimensional sub-surrogate models including $\hat{G}_i(u_i)$, $\hat{G}_{ij}(u_i, u_j)$ and $\hat{G}_{ijk}(u_i, u_j, u_k)$ are obtained by calling $G(u_i, \mathbf{u}_0^i)$, $G(u_i, u_j, \mathbf{u}_0^{ij})$ and $G(u_i, u_j, u_k, \mathbf{u}_0^{ijk})$.

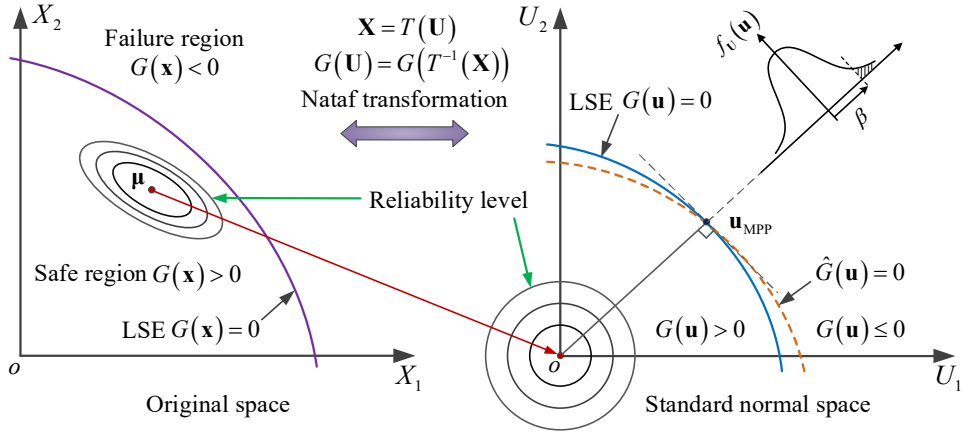

**Fig. 1**. Relationship between original space **X** and standard normal space **U**.

## 3 AL strategy for Kriging-HDMR modeling

The construction of the surrogate model expressed by Eqs. (11) and (12) is a progressive process. Generally, the first-order surrogate models ($\hat{G}_i(u_i)$) have to be included for efficiently reflecting the influence from each random variable. For the second-order and third-order surrogate models ($\hat{G}_{ij}(u_i, u_j)$ and $\hat{G}_{ijk}(u_i, u_j, u_k)$), additional tests are conducted to identify the significance of the coupling effects, thereby the required sub-surrogate models can be determined.

3.1 Surrogate modeling with first-order components

For a LSF with $N_D$ random variables, $N_D$ first-order sub-surrogate models have to be constructed for including the influence of each random variable.

For the $i$-th random variable, three pairs of DoE data, denoted as $\mathbf{S}_{\text{DoE}}^i = (u_{i_0} - \Delta u, u_{i_0}, u_{i_0} + \Delta u)$ and $\mathbf{Y}_{\text{DoE}}^i = \left(G(u_{i_0} - \Delta u, \mathbf{u}_0^i), G(u_{i_0}, \mathbf{u}_0^i), G(u_{i_0} + \Delta u, \mathbf{u}_0^i)\right)$, are initially obtained to build the initial sub-surrogate model, where $\Delta u \in (0, 6.0]$ represents the pre-set quantity. Subsequently, informative samples for the *i*-th sub-surrogate model are obtained step-by-step, and then the *i*-th sub-surrogate model is gradually updated. The optimization mathematical model for finding an informative sample is represented as

$$\begin{aligned} &\text{find} \quad u_i^* \\ &\max \quad F_{\text{obj}}(u_i) = \sigma_{\hat{G}_i}^2(u_i) \\ &\text{s.t.} \quad u_i \in [-u_{\lim}, u_{\lim}] \end{aligned} \quad (16)$$

where $\sigma_{\hat{G}_i}^2(u_i)$ represents the Kriging variance at $u_i$ and $u_i^*$ refers to the selected sample for *i*-th random variable. Once the informative sample $u_i^*$ is selected, the DoE data for *i*-th sub-surrogate model can be updated as $\mathbf{S}_{\text{DoE}}^i = [\mathbf{S}_{\text{DoE}}^i, u_i^*]$ and $\mathbf{Y}_{\text{DoE}}^i = [\mathbf{Y}_{\text{DoE}}^i, G(u_i^*, \mathbf{u}_0^i)]$, and the sub-surrogate model $\hat{G}_i(u_i)$ is updated correspondingly.

For the construction of first-order sub-surrogate models, the stopping criterion is set as

$$\max(\sigma_{\hat{G}_i}(u_i)) < \alpha(\max(\mathbf{Y}_{\text{DoE}}^i) - \min(\mathbf{Y}_{\text{DoE}}^i)) \quad (i = 1, 2, \ldots, N_D) \quad (17)$$

where $\alpha$ is a parameter to control the prediction accuracy.

For better understanding, the LSF $G(u_1, u_2) = 0.75u_2 - 3.0\sin(u_1) + 0.2u_1 - 1.5$ is adopted to demonstrate the construction process of the first-order surrogate models. The cut point is set to $(0,0)$, and then the two functions related to the two first-order sub-surrogate models are $G(u_1, 0) = -3.0\sin(u_1) + 0.2u_1 - 1.5$ and $G(0, u_2) = 0.75u_2 - 1.5$, respectively. In this example, $\mathbf{S}_{\text{DoE}}^1 = (-6.0, 0, 6.0)$ and $\mathbf{S}_{\text{DoE}}^2 = (-6.0, 0, 6.0)$ are initially set to create $\hat{G}_1(u_1)$ and $\hat{G}_2(u_2)$. **Fig. 2**a) shows the added DoE points for updating first-order sub-surrogate models $\hat{G}_1(u_1)$ and the corresponding predicted LSEs, where the numbers indicate the order of acquisition. It can be noted that 5 added DoEs are required to construct the high-accurate sub-surrogate model $\hat{G}_1(u_1)$, while no more added DoE is required for the other first-order sub-surrogate model because the existed three DoEs provide enough information to ensure the accuracy for approximation of the linear function $G(0, u_2)$. **Fig. 2**b) demonstrates the distribution of $\sigma_{\hat{G}_1}^2(u_1)$ before and after adding the second added DoE point shown in **Fig. 2**a). As indicated in **Fig. 2**b), the point with highest Kriging variance (the second added point in **Fig. 2**a)) can be selected, and then the value of Kriging variance near to this point can be significantly reduced when the DoE date of this point is supplemented.

It should be noted that there is no coupling term between variables in this LSF, so the accurate prediction can be achieved by using the first-order sub-surrogate models. For the LSF with coupling effect between variables, further higher-order sub-surrogate models that consider the coupling effect need to be supplemented.

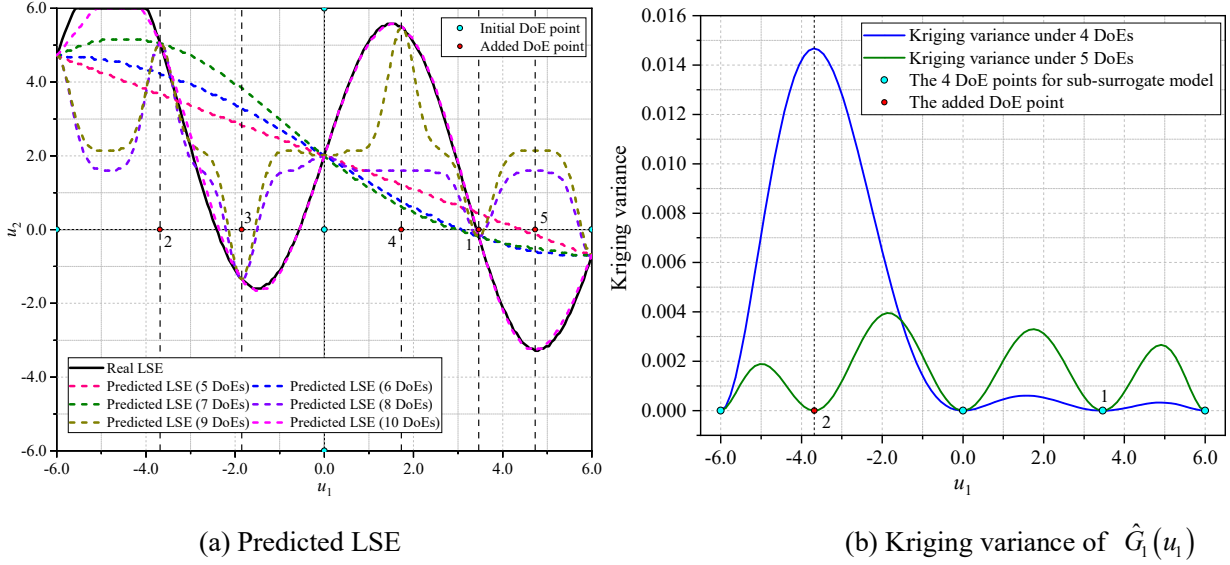

| (a) Predicted LSE | (b) Kriging variance of $\hat{G}_1(u_1)$ |

**Fig. 2**. Construction of surrogate model with first-order components.

### 3.2 Supplement of higher-order components

When significant coupling effects exist among random variables, higher-order sub-surrogate models should be included to improve the accuracy of the overall surrogate model. Typically, second-order sub-surrogate models are firstly supplemented. For more complex coupled systems, third-order sub-surrogate models may be further incorporated if necessary. This section primarily focuses on the method for identifying bivariate coupling effects and establishing the fundamental approach for initializing the second-order sub-surrogate models. Since the identification of trivariate coupling effects and the construction of third-order sub-surrogate models follow analogous procedures, they will not be introduced in detail.

For a given points denoted as $(u_i, u_j, \mathbf{u}_0^{ij})(i<j)$, the significance of the coupling effect can be reflected by the difference between $G(u_i, \mathbf{u}_0^i) + G(u_j, \mathbf{u}_0^j) - G(\mathbf{u}_0)$ and $G(u_i, u_j, \mathbf{u}_0^{ij})$. Since the first-order surrogate models have been constructed by using the method in **Sec. 3.1**, $\hat{G}_i(u_i)$ can be used to predict the value of $G(u_i, \mathbf{u}_0^i)$. Then, by setting the testing point $(u_{i_0} + \Delta u, u_{j_0} + \Delta u, \mathbf{u}_0^{ij})(i<j)$, the significance of the coupling effect between $u_i$ and $u_j$ can be investigated according to the predicted value of $\hat{G}_i(u_{i_0} + \Delta u) + \hat{G}_j(u_{j_0} + \Delta u) - G(\mathbf{u}_0)$ and the LSF value of $G(u_{i_0} + \Delta u, u_{j_0} + \Delta u, \mathbf{u}_0^{ij})$. Specifically, the index $C_{ij}$ defined as follows can be used to evaluate the coupling effect between $u_i$ and $u_j$:

$$C_{ij} = \frac{G(u_{i_0} - \Delta u, u_{j_0} - \Delta u, \mathbf{u}_0^{ij}) - \left[\hat{G}_1(u_{1_0} + \Delta u) + \hat{G}_2(u_{2_0} + \Delta u) - G(\mathbf{u}_0)\right]}{G(u_{i_0} - \Delta u, u_{j_0} - \Delta u, \mathbf{u}_0^{ij})} \tag{18}$$

For a system with $N_D$ random variables, there are totally $N_D(N_D-1)/2$ indices for all combinations between two random variables. Generally, the second-order sub-surrogate models corresponding to the indices with higher values should be prioritized for inclusion, and the number of second-order sub-surrogate models included can be preset in advance. For the coupling effect between $u_i$ and $u_j$ to be included, the second-order sub-surrogate model $\hat{G}_{ij}(u_i, u_j)$ can be initialized by using the DoE data obtained in construction of the first-order sub-surrogate models as

$$\mathbf{S}_{\text{DoE}}^{ij} = \mathbf{S}_{\text{DoE}}^{i} \cup \mathbf{S}_{\text{DoE}}^{j} \tag{19}$$

$$\mathbf{Y}_{\text{DoE}}^{ij} = \mathbf{Y}_{\text{DoE}}^{i} \cup \mathbf{Y}_{\text{DoE}}^{j} \tag{20}$$

For explicit LSFs, the coupling characteristics of variables can usually be predetermined based on the specific expression of the LSF. Therefore, the required coupled sub-surrogate models are well-defined. For implicit LSFs, in the absence of a prior judgment of coupling characteristics, the method presented in this section can be used to determine the coupling effects. Then, higher-order sub-surrogate models are incrementally incorporated during the construction process of the surrogate model. For actual structures, given that the structural form is determined, the coupling characteristics generally do not vary significantly. Consequently, the method introduced in this section can be employed to pre-determine the supplementary coupling sub-surrogate models before conducting reliability analysis, thereby avoiding the repeated identification of coupling characteristics during the modeling process.

3.3 Surrogate model construction with higher-order components

Even if the second-order sub-surrogate models initialized in **Sec. 3.2** are included, there may be significant differences between the surrogate model constructed from $\hat{G}_i(u_i)$ and $\hat{G}_{ij}(u_i, u_j)$ based on Eqs. (11)-(14) and the real LSF, as the current sub-surrogate models have not accurately reflected the corresponding functions. To improve the prediction accuracy, it is important to identify and update the sub-surrogate models with the lowest prediction credibility. The AL strategy is employed to achieve efficient construction of the surrogate model. Construction of the surrogate model are implemented through a series of updating operations, and each updating operation includes the following two operations: the selection of informative sample and the updating of sub-surrogate model.

For the selection of informative samples, the optimization mathematical model is formulated as

$$\begin{aligned}
\text{find} \quad & \mathbf{u}^* = \left(u_1^*, u_2^*, \ldots, u_{N_D}^*\right) \\
\text{min} \quad & F_{\text{obj}}(\mathbf{u}) = \frac{\left|\mu_{\hat{G}}(\mathbf{u})\right| - \delta}{\bar{\sigma}(\mathbf{u})} + p\left[\max\left(\|\mathbf{u}\| - r_c, 0\right) + \max\left(\|\mathbf{u}\| - r_s, 0\right)\right] \\
\text{s.t.} \quad & u_j \in [-u_{\text{lim}}, u_{\text{lim}}] \quad (j = 1, 2, \ldots, N_D)
\end{aligned} \tag{21}$$

where $\mathbf{u} = (u_1, u_2, \ldots, u_{N_D})$ represents a sample of random variables in the standard normal space, which is considered as the design variables, with $N_D$ the number of random variables, and $\mathbf{u}^* = \left(u_1^*, u_2^*, \ldots, u_{N_D}^*\right)$ is the optimum in the range of $u_j \in [-u_{\text{lim}}, u_{\text{lim}}] (j = 1, 2, \ldots, N_D)$. Generally, $u_{\text{lim}} = 6.0$ because the probability of obtaining samples in $(-\infty, -6.0)$ or $(6.0, \infty)$ is extremely small when conducting random sampling according to the standard normal distribution.

The objective function $F_{\text{obj}}(\mathbf{u})$ in Eq. (21) includes two parts. The first term is constructed based on the U-learning function [1] and the small quantity $\delta$ is introduced to describe the offset between the search target and the LSE. In first term, unlike the existing U-learning function [1] that the Kriging variance of the entire surrogate model is used, the quantity that reflects the maximum uncertainty of the selected sub-surrogate models' predictions, denoted as $\bar{\sigma}(\mathbf{u})$, is employed considering that the only the update of sub-surrogate model will be operated, and $\bar{\sigma}(\mathbf{u})$ is expressed as

$$\bar{\sigma}(\mathbf{u}) = \max\left(\max_{1 \leq i \leq N_D}\left(\sigma_{\hat{G}_i}(u_i)\right), \max_{i,j \in C}\left(\sigma_{\hat{G}_{ij}}(u_i, u_j)\right)\right) \tag{22}$$

where $C$ represents the set of selected second-order sub-surrogate models. The second term in $F_{\text{obj}}(\mathbf{u})$ represents the penalty to control the likelihood of selecting informative sample points outside the specified range, considering that the points far away from the origin contribute less to the construction of surrogate model. Specifically, two values of radius ($r_c$ and $r_s$, $r_s \leq r_c$) are set to control this penalty. As depicted in **Fig. 3**, taking the problem with two standard normal variables as an example, by specifying two radius values, the sampling space is partitioned into three regions: the no-penalty region, the single-penalty region, and the double-penalty region. This partitioning is carried out to approximately account for the impact of the probability density distribution on sample representativeness. The penalty coefficient $p$ is set according to the range of LSF $G(\mathbf{u})$ within the sampling space. In this work, the penalty coefficient is set as

$$p = \alpha_s \frac{\overline{G}(\mathbf{u})\big|_{\mathbf{u} \in \text{DoE}} - \underline{G}(\mathbf{u})\big|_{\mathbf{u} \in \text{DoE}}}{4.0} \tag{23}$$

where $\underline{G}(\mathbf{u})\big|_{\mathbf{u} \in \text{DoE}}$ and $\overline{G}(\mathbf{u})\big|_{\mathbf{u} \in \text{DoE}}$ represent the minimum and maximum of the LSF values obtained from the current set of DoE, $\alpha_s = \sqrt{2/N_D}$ refers to the dimensionality adjustment parameter that consider the number of random variables.

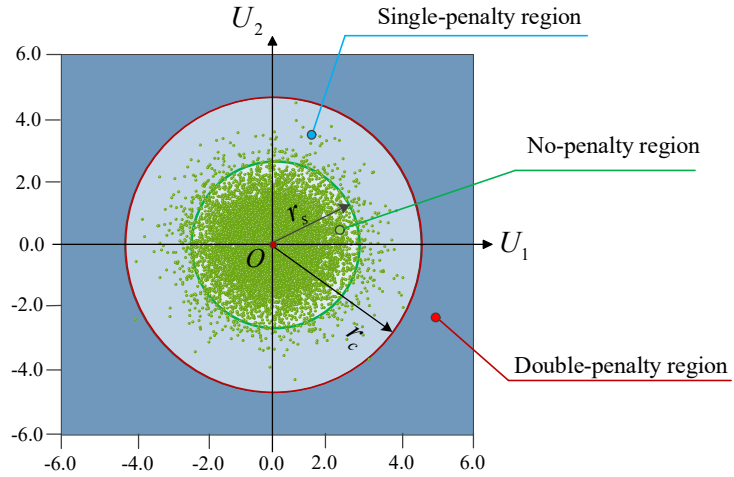

**Fig. 3**. Regional division according to the two radius values.

After obtaining the informative sample encompassing all components, it is necessary to determine the sub-surrogate model for performing the update operation. In this study, the sub-surrogate model with the highest Kriging variance corresponding to the informative sample is selected to be updated. For example, when $\hat{G}_{ij}(u_i^*, u_j^*)$ is the sub-surrogate model with the highest Kriging variance, $\hat{G}_{ij}(u_i, u_j)$ will be updated using $\mathbf{S}_{\text{DoE}}^{ij}$ and $\mathbf{Y}_{\text{DoE}}^{ij}$, which are obtained by

$$\mathbf{S}_{\text{DoE}}^{ij} = \left[ \mathbf{S}_{\text{DoE}}^{ij}, \left(u_i^*, u_j^*\right) \right] \tag{24}$$

$$\mathbf{Y}_{\text{DoE}}^{ij} = \left[ \mathbf{Y}_{\text{DoE}}^{ij}, G\left(u_i^*, u_j^*, \mathbf{u}_0^{ij}\right) \right] \tag{25}$$

Through the above operations, one update of the surrogate model can be achieved. Iterative repetition of the above operations progressively enhances the approximation accuracy of the surrogate model for the real LSE. Refer to [1], the criterion to stop construction of surrogate model is set as

$$\frac{\left|\mu_{\hat{G}_s(\cdot)}\right|}{\sigma_{\hat{G}_s(\cdot)}} \geq 2.0 \tag{26}$$

where $\hat{G}_s(\cdot)$ represents the sub-surrogate model selected to be updated in the current step.

In the context of more intricate coupled systems, it is feasible to incorporate relevant third-order sub-surrogate models on top of the existing second-order coupling factors. The update and enhancement of the surrogate model can continue to be realized following the previously described active learning strategy. Due to space limitations, detailed elaboration on this aspect will not be provided herein.

3.4 Solution of sample selection

In the optimization problems where the random variables serve as design variables, as presented in Eqs. (16) and (21), the objective functions typically take the form of a nonlinear function featuring multiple peaks (or valleys). Consequently, traditional optimization algorithms, such as the gradient-based methods, often converge merely to local optima, thereby generating suboptimal outcomes. By contrast, contemporary optimization algorithms, including GA, ant colony algorithms, and PSO, are better suited for addressing this intricate optimization problem. This is because they have the theoretical potential to identify the global optimum and do not mandate that the optimization function be differentiable or continuous. PSO is an optimization algorithm derived from the foraging behavior of birds [36]. In this algorithm, particles move at random speeds in the solution space and find the optimal solution through information exchange. PSO has been widely applied in fields such as structural optimization [37], structural reliability assessment [11, 38], and reliability-based optimal design [39]. Given the efficacy of PSO in dealing with continuous variable optimization and its straightforward implementation, this paper employs PSO as the principal algorithm for solving the aforementioned optimization problems.

For the optimization problems presented in Eqs. (16) and (21), the particle swarm including $N_{\text{swarm}}$ particles is initially set, and the informative sample is selected through $N_{\text{ite\_max}}$ iteration steps. Generally, each particle has $N_{\text{D}}$ components and the position and velocity for the $i$-th $(i=1,2,\ldots,N_{\text{swarm}})$ particle at the $n$-th $(n=1,2,\ldots,N_{\text{ite\_max}})$ iteration are denoted as $\mathbf{u}_i^{(n)} = \left(u_{i,1}^{(n)}, u_{i,2}^{(n)}, \ldots, u_{i,N_{\text{D}}}^{(n)}\right)$ and $\mathbf{v}_i^{(n)} = \left(v_{i,1}^{(n)}, v_{i,2}^{(n)}, \ldots, v_{i,N_{\text{D}}}^{(n)}\right)$, respectively. Then, the updating of speed and position for the $j$-th $(j=1,2,\ldots,N_{\text{D}})$ component at each iteration can be expressed as

$$v_{i,j}^{(n+1)} = \omega \cdot v_{i,j}^{(n)} + c_1 \cdot rand \cdot \left(u_{pbest,i,j}^{(n)} - u_{i,j}^{(n)}\right) + c_2 \cdot rand \cdot \left(u_{gbest,j}^{(n)} - u_{i,j}^{(n)}\right) \tag{27}$$

$$u_{i,j}^{(n+1)} = u_{i,j}^{(n)} + u_{i,j}^{(n+1)} \tag{28}$$

where $u_{pbest,i,j}^{(n)}$ stands for the $j$-th component of the historical best position of the $i$-th particle at the $n$-th iteration,

$u_{gbest,j}^{(n)}$ represents the *j*-th component of the best position of the entire swarm at the *n*-th iteration. The notation *rand* signifies the generation of a random value uniformly distributed within a specific range of $[0,1]$. The historical best position of a particle and the best position of the entire swarm at the *n*-th iteration are denoted as $\mathbf{u}_{pbest,i}^{(n)} = \left( u_{pbest,i,1}^{(n)}, u_{pbest,i,2}^{(n)}, \ldots, u_{pbest,i,N_D}^{(n)} \right)$ and $\mathbf{u}_{gbest}^{(n)} = \left( u_{gbest,1}^{(n)}, u_{gbest,2}^{(n)}, \ldots, u_{gbest,N_D}^{(n)} \right)$, respectively. The historical best position of a particle $\mathbf{u}_{pbest,i}^{(n)}$ evolves with the iteration process and can be described by

$$\mathbf{u}_{pbest,i}^{(n+1)} = \begin{cases} \mathbf{u}_i^{(n+1)}, & \text{if } F_{fit}\left(\mathbf{u}_i^{(n+1)}\right) > F_{fit}\left(\mathbf{u}_{pbest,i}^{(n)}\right) \\ \mathbf{u}_{pbest,i}^{(n)}, & \text{otherwise} \end{cases} \quad (29)$$

where $F_{fit}(\cdot)$ represents the fitness function, which can be obtained via $F_{fit}(\cdot) = 1/\left(F_{obj}(\cdot) + \delta_0\right)$, where $\delta_0$ is a preset small quantity, such as $1.0 \times 10^{-8}$. In Eq. (27), $\omega$ represents the inertia weight, while $c_1$ and $c_2$ denote the cognition learning factor and social learning factor, respectively. In accordance with the recommendations in existing references [40], these parameters in this paper are set as $\omega = 0.729$ and $c_1 = c_2 = 2.0$. The size of particle swarm and the number of iterations are set to $N_{swarm} = 50$ and $N_{ite\_max} = 50$, respectively. Furthermore, a velocity limit constant $v_{max} = 0.3$ is incorporated to regulate the maximum movement of a particle within each iteration, thus governing the convergence rate and global exploration capability of the algorithm.

Among the historical best positions of all particles, the position with the highest fitness value is designated as the global best position of the entire swarm, which is expressed as

$$\mathbf{u}_{gbest}^{(n)} \in \left\{ \mathbf{u}_{pbest,1}^{(n)}, \mathbf{u}_{pbest,2}^{(n)}, \ldots, \mathbf{u}_{pbest,N_{swarm}}^{(n)} \,\middle|\, F_{fit}\left(\mathbf{u}_{pbest,i}^{(n)}\right) \right\} \quad (30)$$

The key steps of the PSO algorithm employed in this study are outlined below. Firstly, configure the control parameters. Then, randomly initialize particle positions and velocities within the solution space, evaluate each particle, set the current position as the historical best, and determine the global best position via Eq. (30). Subsequently, Update particle positions and velocities using Eqs. (27) and (28) to form a new swarm. Then, reevaluate particles, determine their new historical best positions according to Eq. (29), and update the global best position using Eq. (30). Finally, check if the maximum iteration limit is reached. If so, output the global best position and terminate; otherwise, go back to update particle positions and velocities to form a new swarm. **Fig. 4** provides a comprehensive visualization of the convergence process for acquiring an informative sample point using PSO, documenting the spatial distribution evolution of particles across four key stages: initial configuration, and after 15, 30, and 50 iterative cycles. The directional vectors (rendered in red) illustrate the anticipated movement trajectories of particles during each subsequent iteration phase. The progressive spatial convergence of particles toward a localized region, coupled with the emergence of a stable swarm centroid, demonstrates PSO's efficacy in determining optimal sampling coordinates. This convergence behavior substantiates the algorithm's effectiveness for the proposed sample selection method.

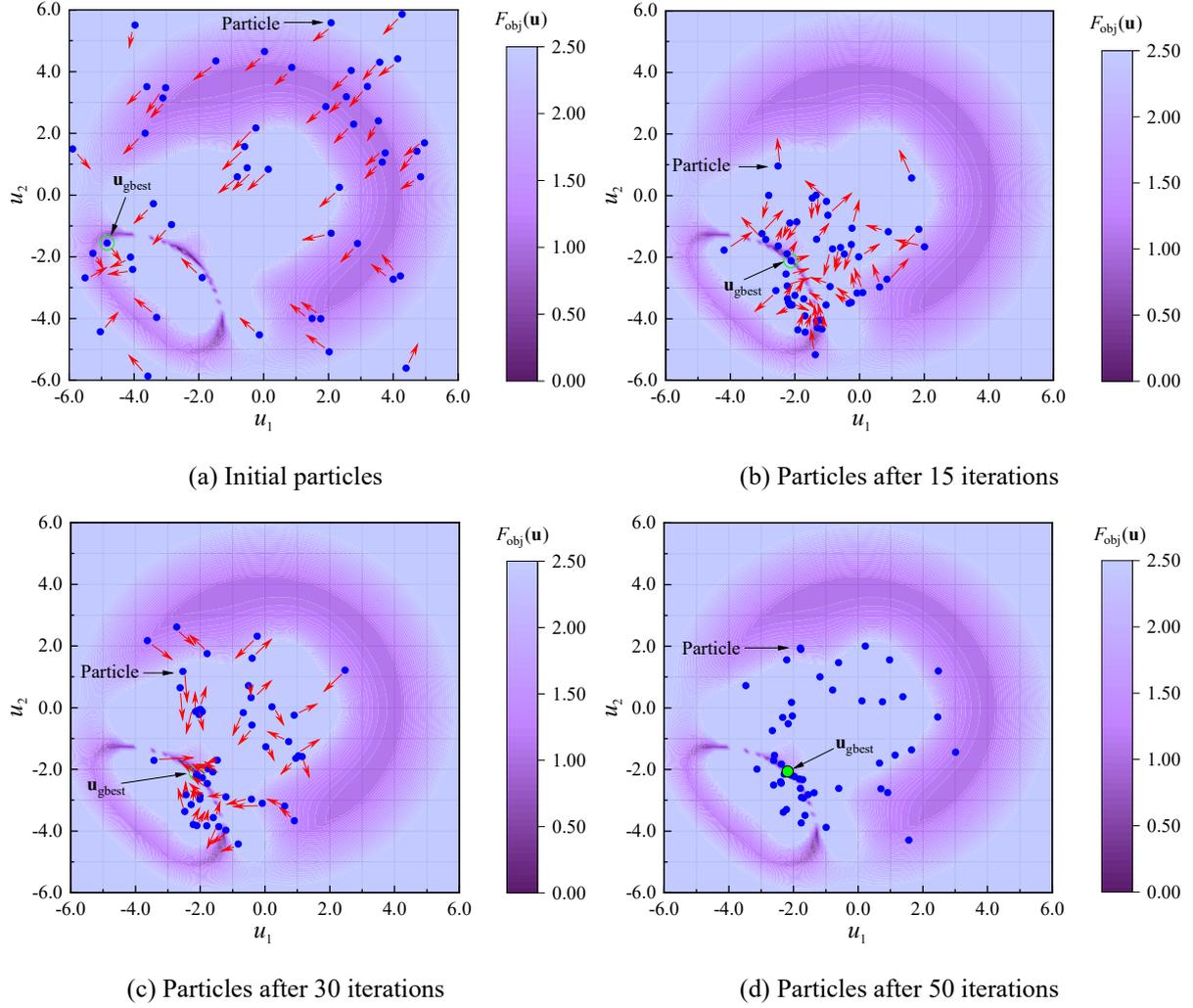

(a) Initial particles

(b) Particles after 15 iterations

(c) Particles after 30 iterations

(d) Particles after 50 iterations

**Fig. 4**. Demonstration of PSO process for the informative sample selection [21].

3.5 Implementation process

    **Fig. 5** demonstrates the implementation process of the proposed AL surrogate model method based on Kriging-HDMR, which includes five stages. The construction of surrogate model is completed through the first four stages, while the **Stage 5** predicts the failure probability through MCS based on the constructed surrogate model. The first four stages respectively accomplish the construction of the first-order sub-surrogate models (**Stage 1**), the selection and initialization of the second-order sub-surrogate models (**Stage 2**), the adaptive updating of the sub-surrogate models (**Stage 3**), and the supplementation of higher-order sub-surrogate models (**Stage 4**). In practical applications, the implementation steps can be simplified according to specific requirements. For example, if the high-order coupled sub-surrogate models have been predetermined, **Stage 2** can be skipped; if third-order and higher-order sub-surrogate models are not considered, **Stage 4** is unnecessary; if coupling effects are completely disregarded, only **Stage 1** needs to be executed, while **Stages 2–4** can be omitted.

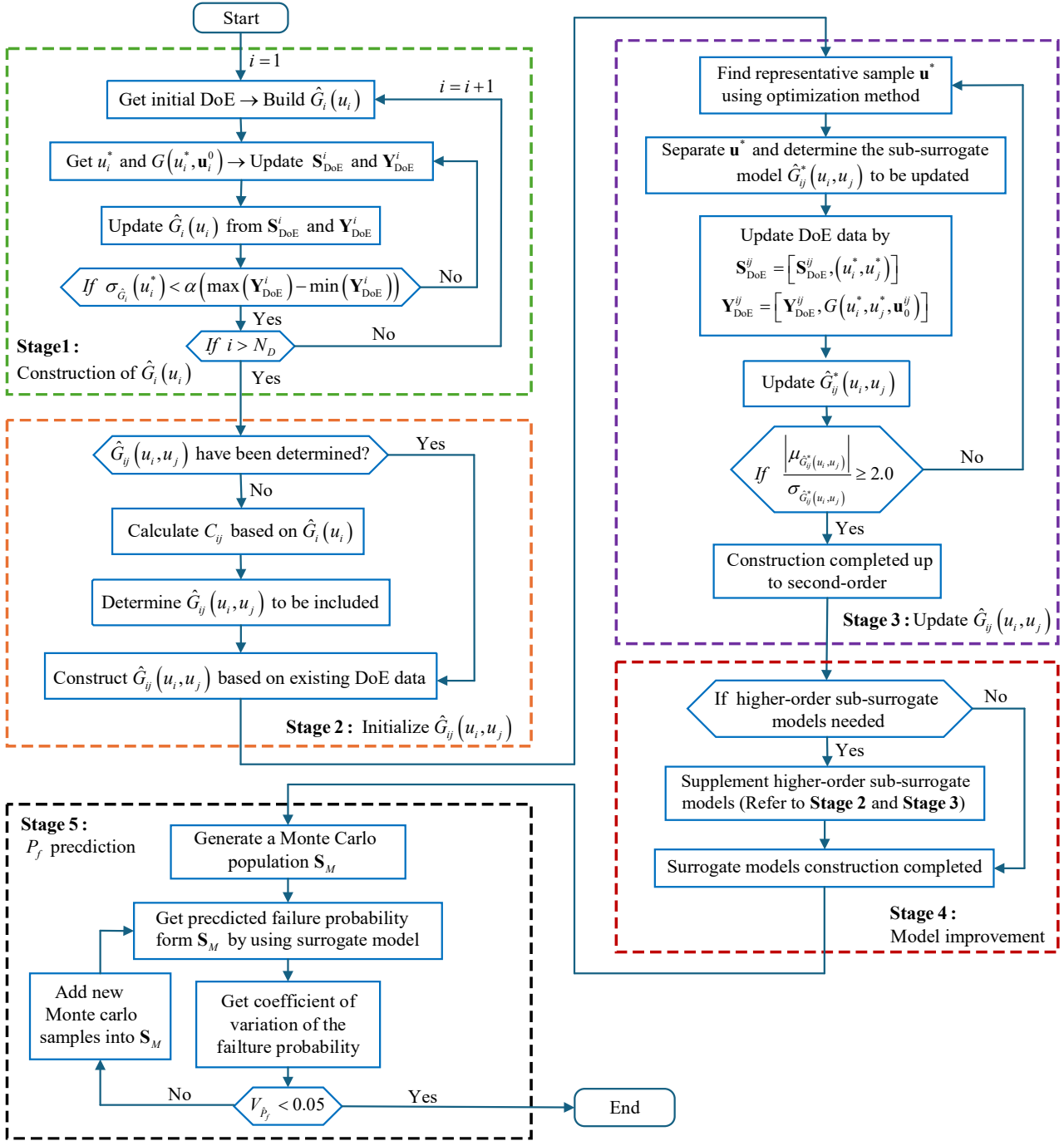

**Fig. 5.** Flow chart of the AL surrogate model method integrating Kriging-HDMR and MCS.

In **Stage 5**, the estimation of failure probability is implemented via MCS based on the constructed surrogate model, and the value of failure probability can be obtained as

$$\hat{P}_f = \frac{n_{\hat{G}<0}}{n_{MC}} \tag{31}$$

where $n_{MC}$ refers to the total number of samples in MCS and $n_{\hat{G}<0}$ represents the number of samples with $\hat{G}<0$ in MCS. The size of Monte Carlo population should be large enough to reduce the impact of sampling randomness on failure probability estimation. Therefore, the variation coefficient of failure probability $V_{\hat{P}_f}$ defined as follows is adopted to evaluate the sufficiency of Monte Carlo population:

$$V_{\hat{P}_f} = \sqrt{\frac{1-\hat{P}_f}{n_{MC}\hat{P}_f}} \tag{32}$$

Generally, the criterion is set to $V_{\hat{P}_f} < 0.05$. For the case with $V_{\hat{P}_f} \geq 0.05$, the Monte Carlo samples should be increased for further estimation of failure probability.

## 4 Validation

In this section, five numerical examples are conducted to access the performance of the proposed method. These examples include: (1) the three-dimensional LSF with second-order coupling items, (2) the high-dimensional LSF with linear first-order terms, (3) the high-dimensional LSF with second-order coupling items, and (4) the truss structure with implicit LSF. For the sake of convenience, the proposed method is denoted as CFAK-H-MCS (CSP-Free Active learning reliability method combining Kriging-HDMR and Monte Carlo Simulation). In the implementation of CFAK-H-MCS, the settings of parameters for the examples are listed in **Table 1**. In addition, the relevant parameters for running of PSO have been listed in **Sec. 3.4** and will not be elaborated here.

**Table 1** Parameter settings for CFAK-H-MCS to the numerical examples.

| Example | $N_D$ | $r_s$ | $r_c$ | $\delta$ | $\alpha$ | Number of $\hat{G}_{ij}(x_i, x_j)$ |
|---|---|---|---|---|---|---|
| 1 | 3 | 2.8 | 3.5 | 0.001 | 0.01 | 2 |
| 2 | 20 | 2.8 | 3.5 | 0.001 | 0.05 | 0 |
|   | 40 | 2.8 | 3.5 | 0.001 | 0.05 | 0 |
|   | 60 | 2.8 | 3.5 | 0.001 | 0.05 | 0 |
|   | 100 | 2.8 | 3.5 | 0.001 | 0.05 | 0 |
| 3 | 20 | 2.8 | 3.5 | 0.001 | 0.05 | 19 |
|   | 60 | 2.8 | 3.5 | 0.001 | 0.05 | 59 |
| 4 | 10 | 2.8 | 3.2 | 0.010 | 0.01 | 10 |
|   | 30 | 2.8 | 3.2 | 0.050 | 0.01 | 30 |

To ensure the credibility of the investigation, multiple runs are executed for the proposed method under specified parameter settings. Subsequently, the following statistical results are employed to evaluate the performance of the proposed method:

(1) The mean value of failure probability prediction $\bar{P}_f$ obtained from $n$ runs, which can be defined as follows

$$\bar{P}_f = \frac{1}{n}\sum_{i=1}^{n}\hat{P}_f^{(i)} \tag{33}$$

where $\hat{P}_f^{(i)}$ represents the failure probability prediction obtained from the $i$-th run.

(2) The mean relative error of failure probability prediction $\bar{\varepsilon}_{\hat{P}_f}$ obtained from $n$ runs, which can be defined as follows

$$\bar{\varepsilon}_{\hat{P}_f} = \frac{1}{n}\sum_{i=1}^{n}\left(\frac{\hat{P}_f^{(i)} - P_{f,\text{MCS}}^{(i)}}{P_{f,\text{MCS}}^{(i)}}\right)\times 100\% \tag{34}$$

where $P_{f,\text{MCS}}^{(i)}$ represents the predicted failure probability obtained from standard MCS in the $i$-th run, $\hat{P}_f^{(i)}$ represents the predicted failure probability obtained from a reliability method in the $i$-th run.

(3) The mean number of calling LSFs for the $n$ runs, which is denoted as $\overline{N}_{\text{call}}$ and obtained by

$$\overline{N}_{\text{call}} = \frac{1}{n}\sum_{i=1}^{n} N_{\text{call}}^{(i)} \tag{35}$$

where $N_{\text{call}}^{(i)}$ represents the number of calls to LSF for the $i$-th run.

4.1 Example 1: three-dimensional LSF with second-order coupling items

In this example, an explicit three-dimensional LSF defined as follows is conducted

$$G(\mathbf{X}) = 0.75x_2 - 3\sin(x_1) + 0.2x_1 - 0.1(x_3 - 3)^2 - 0.005x_1x_2 + 0.1x_2x_3 - 0.2 \tag{36}$$

where $x_1$, $x_2$ and $x_3$ are independent standard normal distributed random variables.

The limit-state function is designed to demonstrate the construction process of surrogates in CFAK-H-MCS. The function consists of three variables and includes seven terms, among which there are two second-order coupling items, four first-order items and one constant term. For CFAK-H-MCS, the cut point is set as $\mathbf{x}_0 = (0, 0, 0)$.

In **Stage 1**, three sub-surrogate models corresponding to the three random variables are constructed. The initial DoE samples for each sub-surrogate model is selected as $\mathbf{S}_{\text{DoE}}^i = (-6.0, 0, 6.0)$ where $i = 1, 2, 3$, and then the sub-surrogate models $\hat{G}_1(x_1)$, $\hat{G}_2(x_2)$ and $\hat{G}_3(x_3)$ are initialized based on these DoE data. Subsequently, the sub-surrogate models $\hat{G}_i(x_i)$ $(i = 1, 2, 3)$ are updated by adding the high informative samples selected through the active learning strategy. **Fig. 6** shows the curves of $G(x_1, 0, 0)$, $G(0, x_2, 0)$ and $G(0, 0, x_3)$ from the LSF and the curves of $\hat{G}_1(x_1)$, $\hat{G}_2(x_2)$ and $\hat{G}_3(x_3)$ obtained by the three sub-surrogate models at the end of this stage, as well as the distribution of the DoE samples corresponding to each sub-surrogate model. As shown in **Fig. 6**, the shape of curves for $G(x_1, 0, 0)$ are more complicate compared with those for $G(0, x_2, 0)$ and $G(0, 0, x_3)$. Correspondingly, the number of DoE samples required for construction of $\hat{G}_1(x_1)$, which is 11, is significantly higher than those for the construction of $\hat{G}_3(x_3)$ and $\hat{G}_2(x_2)$.

For the present LSF, two second-order sub-surrogate models need to be created. In **Stage 2**, $\Delta x$ is set to 2.0 and then the three indices for evaluation of the coupling effect are work out as $C_{12} = 0.0248$, $C_{13} = 0.00$ and $C_{23} = 0.2638$. It can be realized that the coupling effect between $(x_1, x_2)$ and $(x_2, x_3)$ cannot be neglected according to the values of $C_{12}$ and $C_{23}$. Therefore, the sub-surrogated models $\hat{G}_{12}(x_1, x_2)$ and $\hat{G}_{23}(x_2, x_3)$ should be introduced. In the implementation, the DoE data from **Stage 1**, such as $\text{DoE}_1 \cup \text{DoE}_2$ and $\text{DoE}_2 \cup \text{DoE}_3$, can be used to create the two sub-surrogate models $\hat{G}_{12}(x_1, x_2)$ and $\hat{G}_{23}(x_2, x_3)$, respectively.

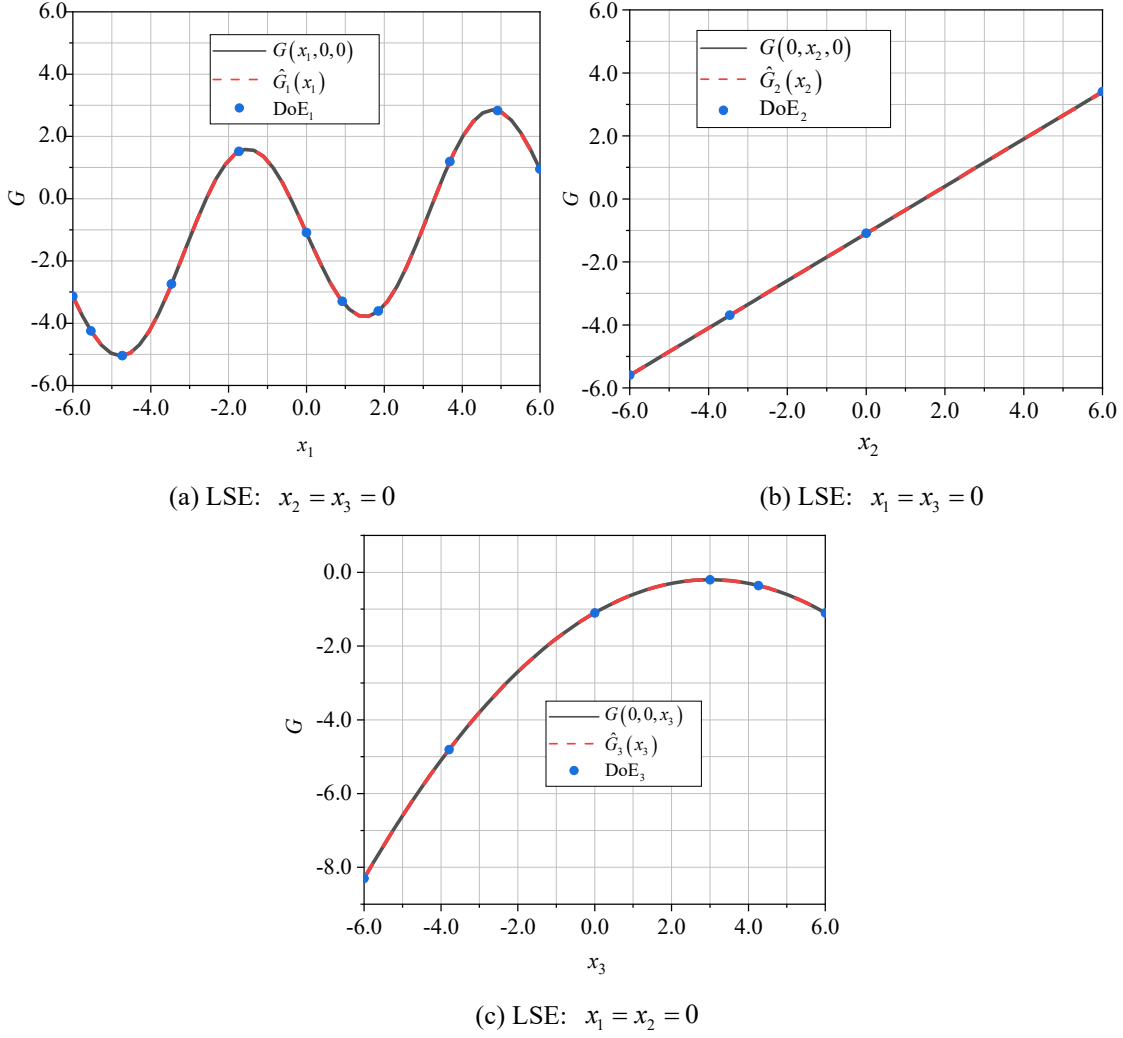

**Fig. 6** Evaluation of first-order sub-surrogate models (Example 1).

In **Stage 3**, the active learning strategy is operated to improve the sub-surrogate models, especially for the second-order sub-surrogate models, until the stop criterion is met. **Fig. 7** shows the real LSEs under the settings of $x_1 = 0$, $x_2 = 0$ and $x_3 = 0$ as well as the predicted LSEs obtained by the surrogate models corresponding to different sets of DoE data. The pink dashed lines in **Fig. 7** represents the predicted LSE curves using the initial $\hat{G}_{12}(x_1, x_2)$ and $\hat{G}_{23}(x_2, x_3)$ constructed from the DoE data in **Stage 1**. It can be seen that, in the beginning of **Stage 3**, there is a significant deviation between the predicted LSEs and the real LSEs in **Fig. 7**a) and **Fig. 7**b). Through the update process of surrogate models, the final predicted LSE of $\hat{G}_{23}(x_2, x_3)$ constructed with 10 DoE samples completely coincides with the real LSE, as shown in **Fig. 7**b), and the final predicted LSE of $\hat{G}_{23}(x_2, x_3)$ constructed with 24 DoE samples provides accurate prediction in the region inside the control circle (see **Fig. 7**a)), due to the complexity of $G(x_1, x_2, 0)$. Under the setting of $x_2 = 0$, the accurate LSE can be obtained as shown in **Fig. 7**c) at the beginning of **Stage 3**, because the predicted LSE is actually composed of $\hat{G}_1(u_1)$ and $\hat{G}_2(u_2)$, without coupling effect between $x_1$ and $x_3$.

To better illustrate the process of updating surrogate models in **Stage 3**, a comparison between the predicted LSE of $\hat{G}_{12}(x_1, x_2)$ and the real LSE along with the increase of DoE samples is presented in **Fig. 8**. Under the constraint of the circles with preset parameters $r_c = 3.5$ and $r_s = 2.8$, high informative DoE samples are selected based on the optimization mathematical model and the active learning strategy. It can be seen that the predicted LSEs obtained by $\hat{G}_{12}(x_1, x_2)$ approaches the real LSE gradually, and the final predicted LSE basically matches the ture LSE as shown in **Fig. 8**d). Although the final predicted LSE does not perfectly match the real LSE, it is ensured that within the circles, the predicted LSE is almost completely coincident with the real LSE. Since the random variables in this example follow the standard normal distribution, the main area of the variable distribution is covered within the outer circle. Therefore, the accuracy of failure probability prediction is determined by the predicted LSE inside this area.

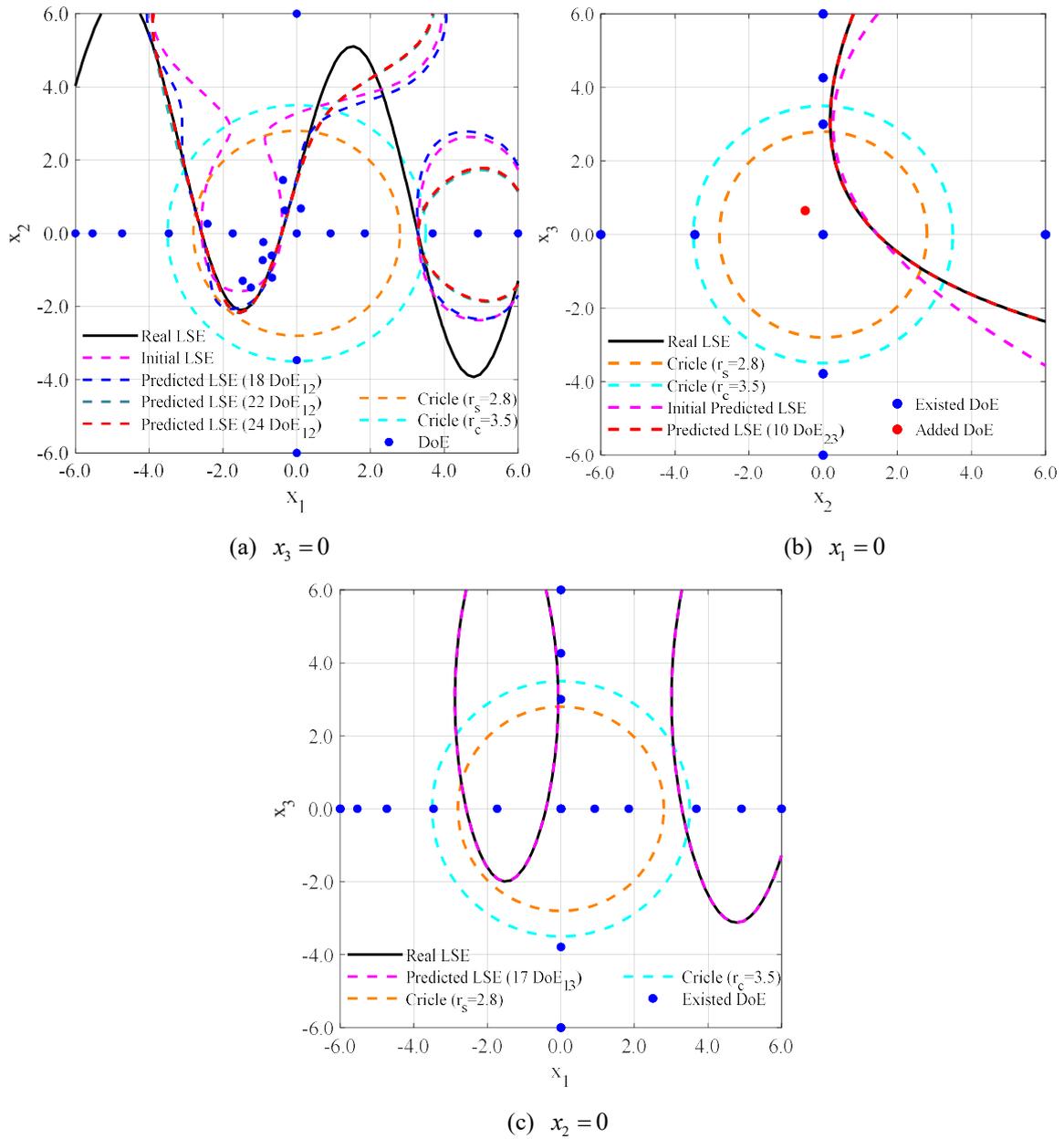

(a) $x_3 = 0$        (b) $x_1 = 0$

(c) $x_2 = 0$

**Fig. 7** Comparison of real LSEs and predicted LSEs from second-order sub-surrogate models (Example 1).

For the present LSF, $C_{123} = 0$ because there is no third-order coupling item. Therefore, there is no need to construct the third-order sub-surrogate models in **Stage 4**. Thus, it enters **Stage 5** for failure probability estimation. Under the setting of Monte Carlo population as $1\times10^6$, the predicted failure probability obtained by the proposed method is 0.6834%, and the relative error compared to the failure probability obtained by MCS, which is 0.6835%, is only 0.0132%. Note that only 32 calls of LSF are required in the proposed method.

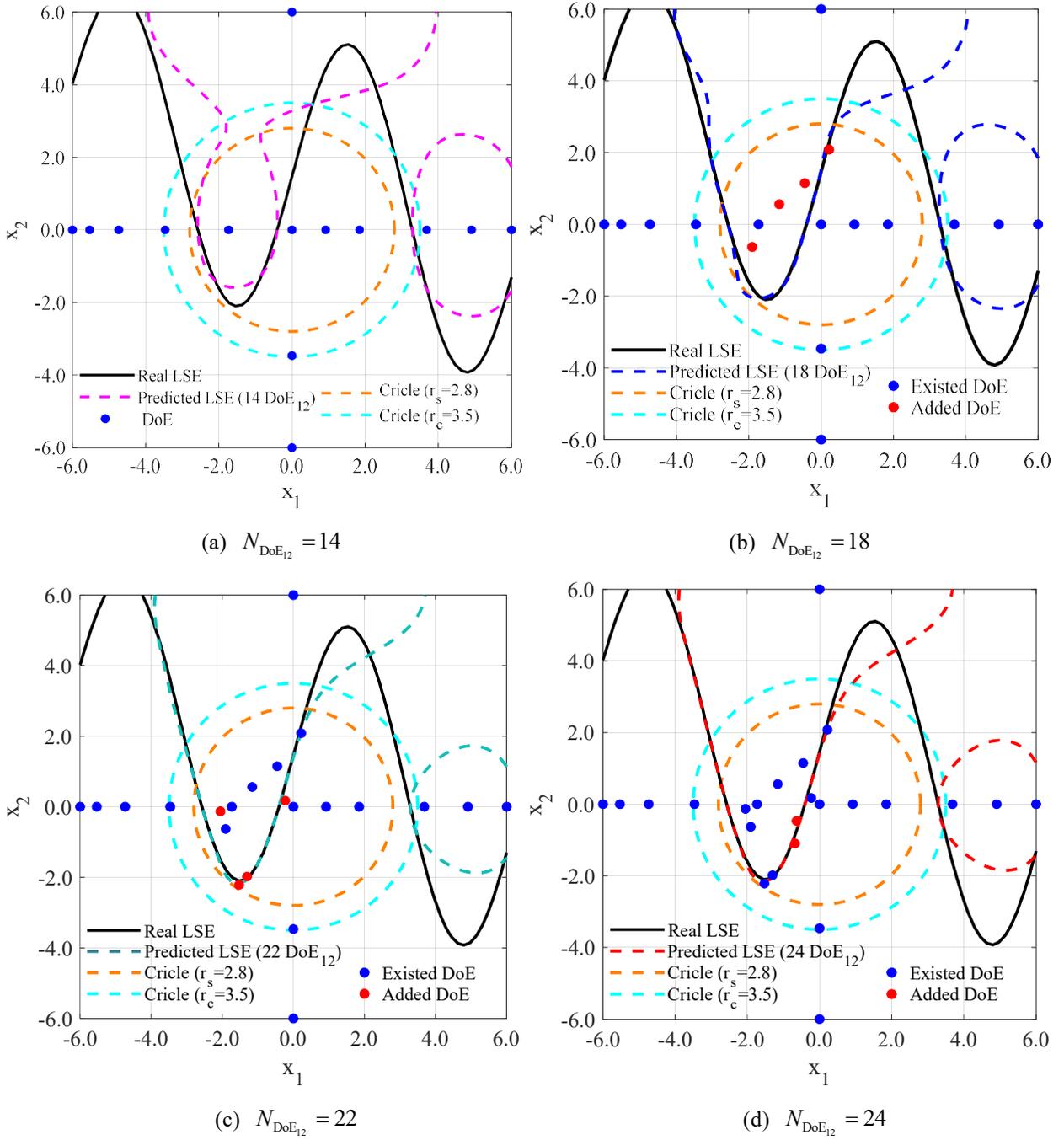

**Fig. 8** Improvement process of $\hat{G}_{12}(x_1, x_2)$ (Example 1).

4.2 Example 2: high-dimensional LSF with linear first-order terms

The explicit high-dimensional LSF with linear first-order terms is defined as [41]

$$G(\mathbf{x}) = 3.5\sqrt{N_D} - \sum_{i=1}^{N_D} x_i \tag{37}$$

where $x_i (i=1,2,\ldots,N_D)$ are independent standard normal distributed random variables, and $N_D$ represents the total number of random variables. Note that the exact failure probability is $P_f = \Phi(-3.5)$, regardless of the problem dimension.

In estimation of the failure probability, Monte Carlo population is set to $2\times10^6$. In implementation of CFAK-H-MCS, $\Delta u$ is set to 6.0, and the cut point $u_{i_0}$ for each random variable is taken as 0. Then, the initial DoE sets for the first-order sub-surrogate models are $\mathbf{S}_{\text{DoE}}^i = (-6.0, 0, 6.0)$, where $i=1,2,\ldots,N_D$. According to the expression of the LSF, no second-order coupling sub-surrogate model is required. In other words, the algorithm only runs two stages: **Stage 1** and **Stage 5**. For various methods, multiple runs are performed to ensure the credibility of the algorithm performance evaluation. In this investigation, four settings of $N_D$ including $N_D = 20$, $N_D = 40$, $N_D = 60$ and $N_D = 100$ are considered, and the statistical results for the 30 runs of MCS, AK-MCS+U [1] and CFAK-H-MCS are presented in **Table 2**, where the results of AK-MCS+U are provided by [21].

**Table 2** Results of the high-dimensional LSF with linear first-order terms.

| $N_D$ | Method | $\bar{N}_{\text{call}}$ | $\bar{P}_f$ | $\bar{\varepsilon}_{\hat{P}_f}$ / % | $\bar{V}_{\hat{P}_f}$ / % |
|---|---|---|---|---|---|
| 20 | MCS | $2\times10^6$ | $2.307\times10^{-4}$ | 0.56 | 4.658 |
|  | AK-MCS+U | 74.9 | $2.293\times10^{-4}$ | 1.16 | 3.504 |
|  | CFAK-H-MCS | 61.6 | $2.307\times10^{-4}$ | 0.56 | 4.658 |
| 40 | MCS | $2\times10^6$ | $2.339\times10^{-4}$ | 0.82 | 4.606 |
|  | AK-MCS+U | 131.1 | $2.292\times10^{-4}$ | 1.21 | 3.504 |
|  | CFAK-H-MCS | 121.1 | $2.339\times10^{-4}$ | 0.82 | 4.606 |
| 60 | MCS | $2.0\times10^6$ | $2.326\times10^{-4}$ | 0.26 | 4.638 |
|  | AK-MCS+U | 220.7 | $2.278\times10^{-4}$ | 1.81 | 3.506 |
|  | CFAK-H-MCS | 181.3 | $2.326\times10^{-4}$ | 0.26 | 4.638 |
| 100 | MCS | $2\times10^6$ | $2.345\times10^{-4}$ | 1.08 | 4.620 |
|  | AK-MCS+U | 361.2 | $2.241\times10^{-4}$ | 3.41 | 3.510 |
|  | CFAK-H-MCS | 301.1 | $2.345\times10^{-4}$ | 1.08 | 4.620 |

Note: The reference failure probability is $P_{f,\text{ref}} = \Phi(-3.5) = 2.32\times10^{-4}$.

As shown in **Table 2**, the number of DoE samples required by CFAK-H-MCS is significantly less than those required by AK-MCS+U, indicating the advance in efficiency of the proposed method. Moreover, the predicted values of failure probability obtained by the proposed method are very close to the results of MCS, which shows the superiority of CFAK-H-MCS in solution accuracy. According to the listed results, it can be observed that the number of DoE samples required by the proposed method roughly shows a linear growth trend as the number of random variables increases. In other words, for the high-dimensional LSF with linear terms only, the introduction of HDMR can effectively avoid the problem of dimension explosion. For the four settings of $N_D$, the boxplots of failure probability obtained by CFAK-H-MCS and AK-MCS+U are presented in **Fig. 9**, with the red dashed line showing as the reference values obtained by MCS. **Fig. 9** illustrates that the stability of CFAK-H-MCS is significantly higher than that of AK-MCS+U.

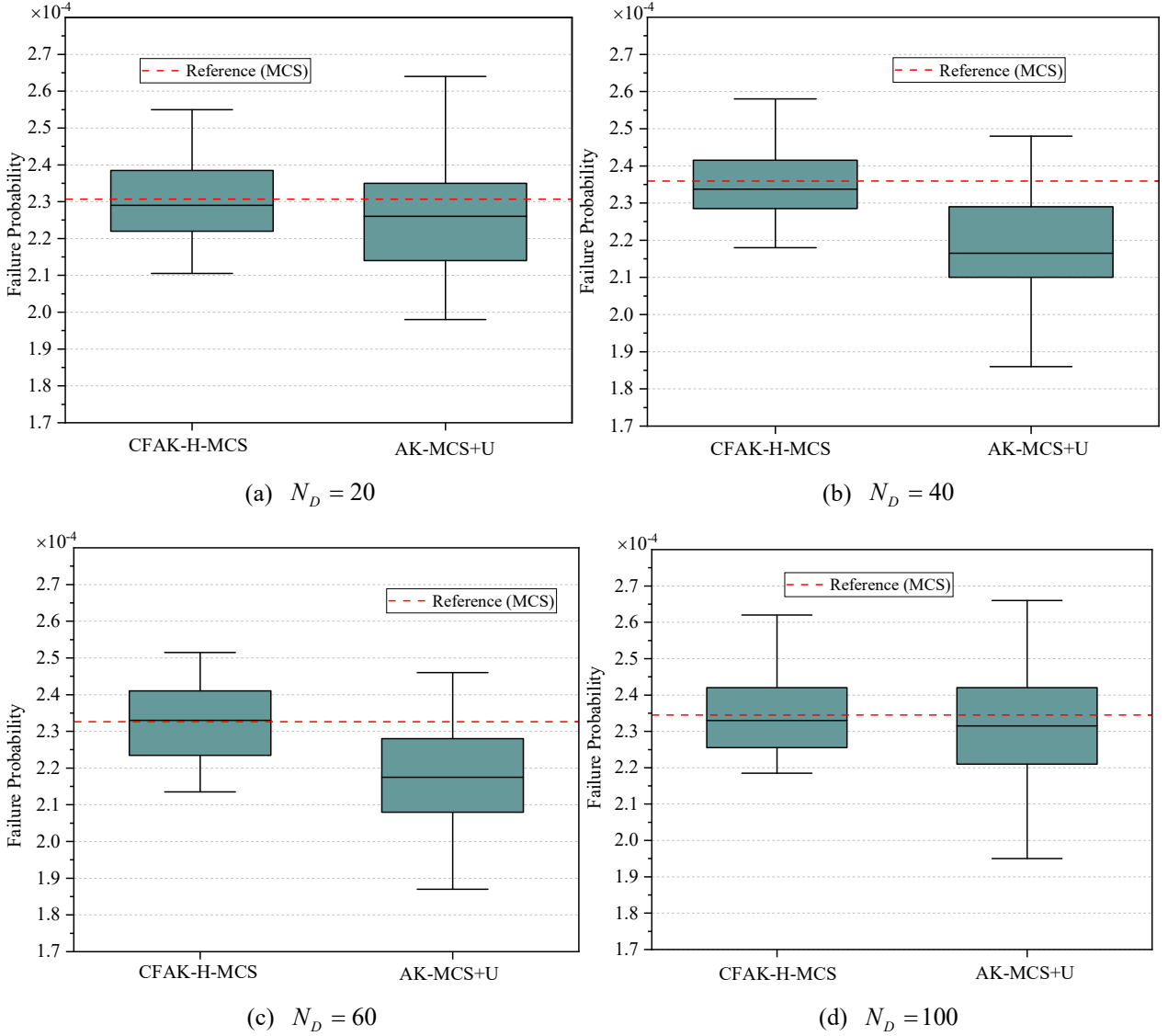

**Fig. 9**. Boxplots of predicted failure probability obtained by CFAK-H-MCS and AK-MCS+U.

4.3 Example 3: high-dimensional LSF with second-order coupling items

The explicit high-dimensional LSF with second-order coupling items is defined as [5]

$$G(\mathbf{x}) = a - (x_1 - 1)^2 - \sum_{i=2}^{N_D} i\left(2x_i^2 - x_{i-1}\right)^2 \tag{38}$$

where $x_i \, (i = 1, 2, \ldots, N_D)$ are independent Gaussian distribution random variables with the mean value of 3.41 and the standard deviation of 0.2, and $N_D$ is the total number of random variables. In Eq. (38), $a$ is a constant. For the cases of $N_D = 20$ and $N_D = 60$, which are considered in the present investigation, the values of $a$ are given as 95000 and 830000, respectively.

In estimation of the failure probability, Monte Carlo population is set to $1\times10^6$. Be similar to Examples 2 and 3, the initial DoE sets for construction of all first-order sub-surrogate models in CFAK-H-MCS are $\mathbf{S}_{\text{DoE}}^i = (-6.0, 0, 6.0)$, where $i = 1, 2, \ldots, N_D$. According to the expression of LSF, the required second-order sub-surrogate models $\hat{G}_{ij}(u_i, u_j)$ can be directly determined without executing **Stage 2**. The results of the proposed method compared with other methods are listed in **Table 3**, where the data of CFAK-H-MCS and MCS presents the statistical results of 50 independent runs and the results of FORM, AK-MCS and AAE-SDR are provided by [5].

**Table 3** demonstrate that the accuracy and efficiency of CFAK-H-MCS are significantly higher than those of traditional methods such as FORM and AK-MCS [1]. Compared with the existing efficient reliability methods such as AAE-SDR [5] and HDDA [42], CFAK-H-MCS still shows higher computational performance. The results verify the good performance of the proposed method in dealing with high-dimensional reliability problems with second-order coupling items. **Fig. 10** shows the boxplots of predicted failure probability and relative error obtained by CFAK-H-MCS. It can be seen that, for the case of $N_D = 20$, the predicted failure probabilities have a 99.3% probability of being in the interval of $[3.39, 3.46]\times10^{-2}$, and the relative errors have a 99.3% probability of being in the interval of $[0, 0.0049]$. For the case of $N_D = 60$, the predicted failure probabilities have a 99.3% probability of being in the interval of $[9.75, 9.86]\times10^{-4}$, and the relative errors have a 99.3% probability of being in the interval of $[0.0128, 0.0237]$. For the first 5 runs, **Fig. 11** shows the change of the predicted failure probability of CFAK-H-MCS along with the increase of DoE samples. For the cases of $N_D = 20$ and $N_D = 60$, the predicted failure probabilities remain zero before 41 DoE samples and 121 DoE samples, respectively. Subsequently, the predicted failure probabilities gradually approach the reference values obtained by MCS, and finally get close to the accurate values of failure probability. These results show that CFAK-H-MCS has good stability in dealing with the reliability problem of high-dimensional explicit LSF.

**Table 3** Results of the high-dimensional LSF with second-order coupling items.

| $N_D$ | Method | $\overline{N}_{\text{call}}$ | $\overline{P}_f$ | $\overline{\varepsilon}_{\hat{P}_f}$ / % | $\overline{V}_{\hat{P}_f}$ / % |
|---|---|---|---|---|---|
| 20 | MCS | $1\times10^6$ | $3.414\times10^{-2}$ | - | 0.55 |
| | FORM | 126 | $1.38\times10^{-2}$ | 59.5 | 0.84 |
| | AK-MCS | 620 | $3.44\times10^{-2}$ | 0.90 | 0.52 |
| | HDDA | 157 | $4.31\times10^{-2}$ | 26.4 | 0.47 |
| | AAE-SDR | 148 | $3.38\times10^{-2}$ | 1.00 | 0.53 |
| | CFAK-H-MCS | 90.3 | $3.418\times10^{-2}$ | 0.18 | 0.53 |
| 60 | MCS | $1\times10^6$ | $9.63\times10^{-4}$ | - | 3.22 |
| | FORM | 372 | $7.98\times10^{-4}$ | 91.7 | 3.53 |
| | AK-MCS | 1020 | $9.97\times10^{-4}$ | 3.50 | 3.16 |
| | AAE-SDR | 380 | $9.26\times10^{-4}$ | 3.40 | 3.28 |
| | CFAK-H-MCS | 270.7 | $9.80\times10^{-4}$ | 1.81 | 3.19 |

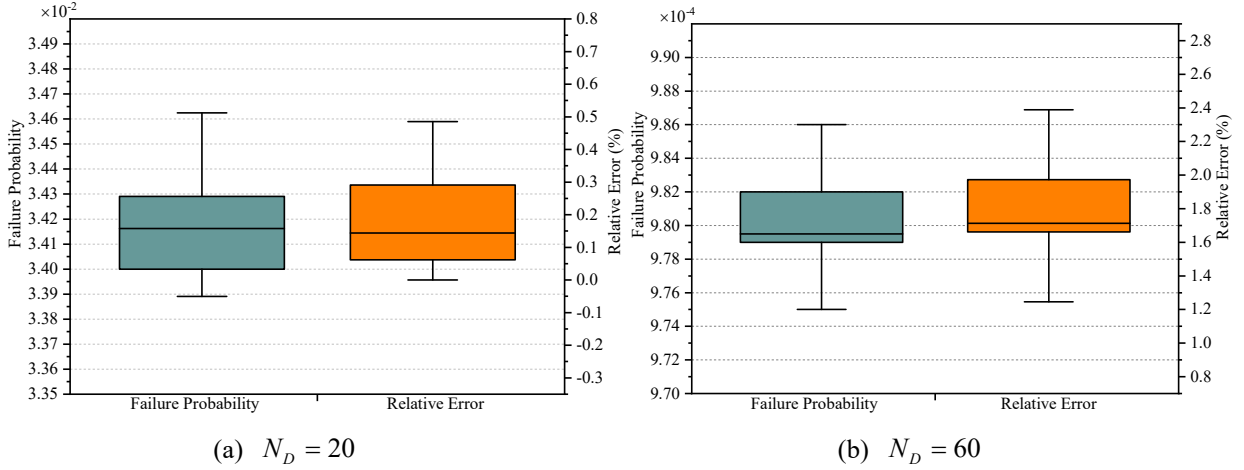

(a) $N_D = 20$      (b) $N_D = 60$

**Fig. 10**. Boxplots of predicted failure probability and relative error (CFAK-H-MCS).

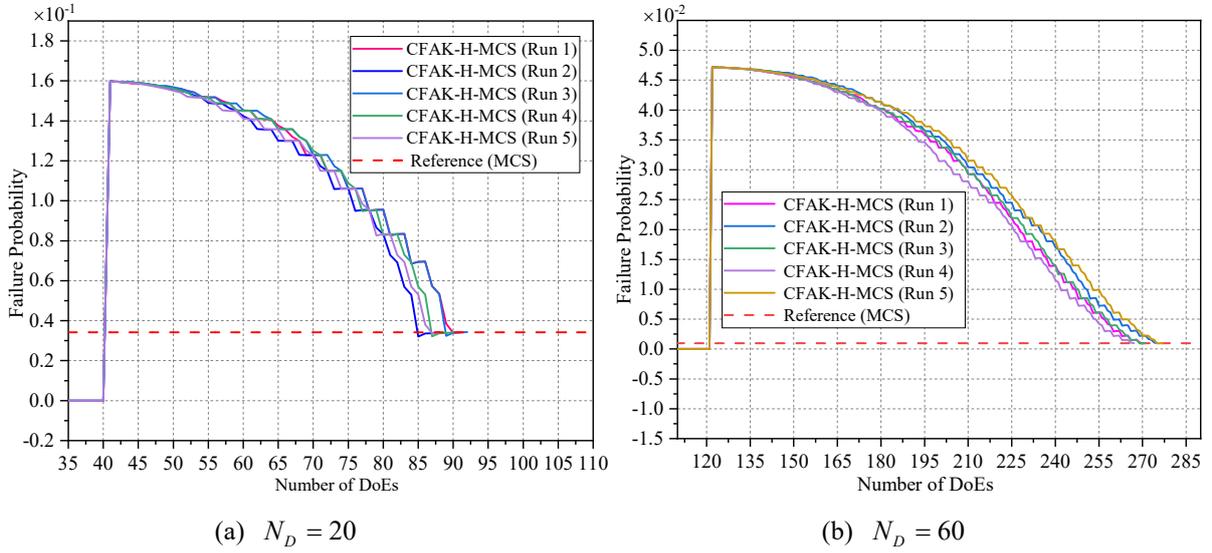

(a) $N_D = 20$      (b) $N_D = 60$

**Fig. 11**. Predicted failure probability along with the number of DoE samples (CFAK-H-MCS).

4.5 Example 4: truss structure with implicit LSF

The geometry and loads of the two-dimensional truss structure are shown in **Fig. 12**, and the implicit LSF is defined as [21]

$$G(\mathbf{x}) = 0.11 - |\Delta(\mathbf{x})| \tag{39}$$

where $\Delta(\mathbf{x})$ represents the vertical displacement at the mid-span, as shown in **Fig. 12**.

Two settings of random variables including the set with 10 random variables and the set with 30 random variables are considered in this investigation. Specifically, the physical meaning, the form of probability distribution form, the mean value and the standard deviation of each random variable are given in **Table 4** and **Table 5**, respectively for the setting of 10 random variables and the 30 random variables. Especially, the subscripts in cross-sectional area $A_i$ and elastic modulus $E_i$ in **Table 4** and **Table 5** represent the indices of members, which have been marked in **Fig. 12**.

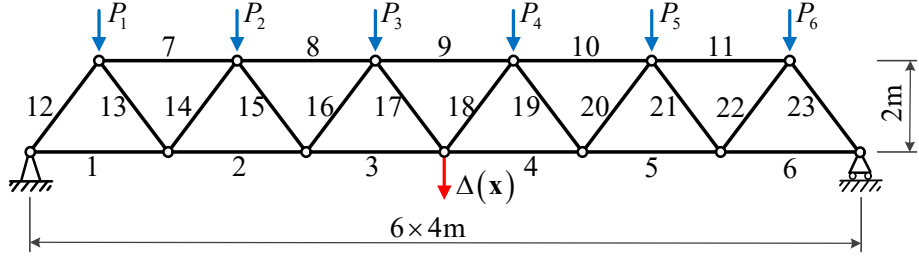

**Fig. 12.** The diagram of structural member numbering.

In estimation of failure probability, the Monte Carlo population is set to $1.0\times10^6$. In implementation of CFAK-H-MCS, $\Delta u$ is set to 3.0, and the cut point $u_{i_0}$ of each random variable is taken as 0. Then, the initial DoE samples for construction of all first-order sub-surrogate models are $\mathbf{S}_{\text{DoE}}^i = (-3.0,\ 0,\ 3.0)$, where $i=1,2,\ldots,N_D$. For the two settings with 10 random variables and 30 random variables, the results of the proposed method and the other methods are listed in **Table 6** and **Table 7**, respectively. Especially, the data of CFAK-H-MCS, MCS, AK-MCS+U [1] and CFAK-MCS [21] are obtained according to the results of 50 runs. In the case of 30 random variables, to avoid large errors due to premature termination, the minimum number of DoE updates for the surrogate model construction in AK-MCS+U and CFAK-MCS is set to 1200.

**Table 4** Details for the setting of 10 random variables.

| Random variable | Physics meaning | Distribution | Mean | Standard deviation |
| --- | --- | --- | --- | --- |
| $x_1$ | $A_1$-$A_{11}$ (m$^2$) | Lognormal | $2.0\times10^{-3}$ | $2.0\times10^{-4}$ |
| $x_2$ | $A_{12}$-$A_{23}$ (m$^2$) | Lognormal | $1.0\times10^{-3}$ | $1.0\times10^{-4}$ |
| $x_3$ | $E_1$-$E_{11}$ (Pa) | Lognormal | $2.1\times10^{11}$ | $2.1\times10^{10}$ |
| $x_4$ | $E_{12}$-$E_{23}$ (Pa) | Lognormal | $2.1\times10^{11}$ | $2.1\times10^{10}$ |
| $x_5$-$x_{10}$ | $P_1$-$P_6$ (N) | Lognormal | $5.0\times10^4$ | $7.5\times10^3$ |

**Table 5** Details for the settings of 30 random variables.

| Random variable | Physics meaning | Distribution | Mean | Standard deviation |
| --- | --- | --- | --- | --- |
| $x_1$-$x_{11}$ | $A_1$-$A_{11}$ (m$^2$) | Lognormal | $2.0\times10^{-3}$ | $2.0\times10^{-4}$ |
| $x_{12}$-$x_{23}$ | $A_{12}$-$A_{23}$ (m$^2$) | Lognormal | $1.0\times10^{-3}$ | $1.0\times10^{-4}$ |
| $x_{24}$ | $E_1$-$E_{23}$ (Pa) | Lognormal | $2.1\times10^{11}$ | $2.1\times10^{10}$ |
| $x_{25}$-$x_{30}$ | $P_1$-$P_6$ (N) | Lognormal | $5.0\times10^4$ | $7.5\times10^3$ |

The results indicate that, for the setting of 10 random variables, CFAK-H-MCS exhibits significant advantages in computational efficiency and accuracy compared to IS, PCE and AK-MCS+U. Since the number of random variables is still relatively low, CFAK-MCS based on optimization to obtain high informative samples demonstrates higher performance. It should be noted that, as the number of random variables increases, the performance of the surrogate model method will be severely challenged. For the setting of 30 random variables, the performance of AK-MCS+U and CFAK-MCS significantly decreases, while CFAK-H-MCS exhibits significant advantages. According to the results as shown in **Table 7**, the number of DoE samples required for CFAK-H-MCS to construct the surrogate models is 298.1, which is significantly lower than those required by AK-MCS+U and CFAK-MCS. More importantly, under the premise of requiring fewer DoE samples, the accuracy of predicted failure probability obtained by CFAK-H-MCS is significantly higher than that of AK-MCS+U and CFAK-MCS. These results demonstrate the good performance of the proposed method in handling reliability problems with high-dimensional implicit LSF. It should be noted that for CFAK-H-MCS, $\overline{N}_{\text{call}}$ in **Table 6** and **Table 7** is expressed as the sum of the number of DoE samples and the number of additional LSF calls for identification of the coupling effects between two random variables (on the right side of the plus sign), the data of which are not included as DoE data to construct the surrogate models. For practical problems, if important coupling effects between two random variables can be determined through other effective methods, these calls of LSF can be saved.

**Table 6** Results of the reliability analysis for the truss structure with 10 random variables.

| Method | $\overline{N}_{\text{call}}$ | $\overline{P}_f$ | $\overline{\varepsilon}_{\hat{P}_f}$ / % | $\overline{V}_{\hat{P}_f}$ / % |
|---|---|---|---|---|
| MCS | $1.0 \times 10^6$ | $8.875 \times 10^{-3}$ | - | 1.06 |
| IS [43] | $2.0 \times 10^5$ | $8.886 \times 10^{-3}$ | 0.10 | - |
| Full PCE [43] | 443 | $8.866 \times 10^{-3}$ | 0.10 | - |
| Sparse PCE [43] | 207 | $8.866 \times 10^{-3}$ | 0.10 | - |
| AK-MCS+U | 227.1 | $7.914 \times 10^{-3}$ | 11.1 | 1.12 |
| CFAK-MCS [21] | 84.9 | $8.923 \times 10^{-3}$ | 0.54 | 1.05 |
| CFAK-H-MCS | 101.3+45 | $8.836 \times 10^{-3}$ | 1.51 | 1.06 |

**Table 7** Results of the reliability analysis for the truss structure with 30 random variables.

| Method | $\overline{N}_{\text{call}}$ | $\overline{P}_f$ | $\overline{\varepsilon}_{\hat{P}_f}$ / % | $\overline{V}_{\hat{P}_f}$ / % |
|---|---|---|---|---|
| MCS | $1.0 \times 10^6$ | $4.941 \times 10^{-3}$ | - | 1.41 |
| AK-MCS+U | 1235.2 | $4.567 \times 10^{-3}$ | 9.67 | 1.47 |
| CFAK-MCS | 1315.1 | $4.723 \times 10^{-3}$ | 5.41 | 1.45 |
| CFAK-H-MCS | 298.1+435 | $4.910 \times 10^{-3}$ | 1.91 | 1.42 |

For the two settings of random variables, **Fig. 13** shows the boxplots of predicted failure probability and relative error obtained by CFAK-H-MCS. It can be seen that, under setting with 10 random variables, the failure probability predictions have a 99.3% probability of falling within the range of $[8.5, 9.2] \times 10^{-3}$, and the relative errors have a 99.3% probability of falling within the range of [0, 0.0092]. For the setting with 30 random variables, the failure probability predictions have a 99.3% probability of falling within the range of approximately $[4.65, 4.90] \times 10^{-4}$, and the relative errors have a 99.3% probability of falling within the range of [0, 0.06]. **Fig. 14** demonstrates the change of the predicted failure probability along with the increase of DoE samples (for the first 5 runs). For both settings with 10 and 30 random variables, the failure probability is estimated to be 0 when the number of DoE samples is lower than 65 and 245, respectively. Subsequently, the failure probability predictions gradually approach the reference values obtained by MCS as the DoE increases, ultimately yield accurate values of failure probability. These results demonstrate the good stability of the proposed method.

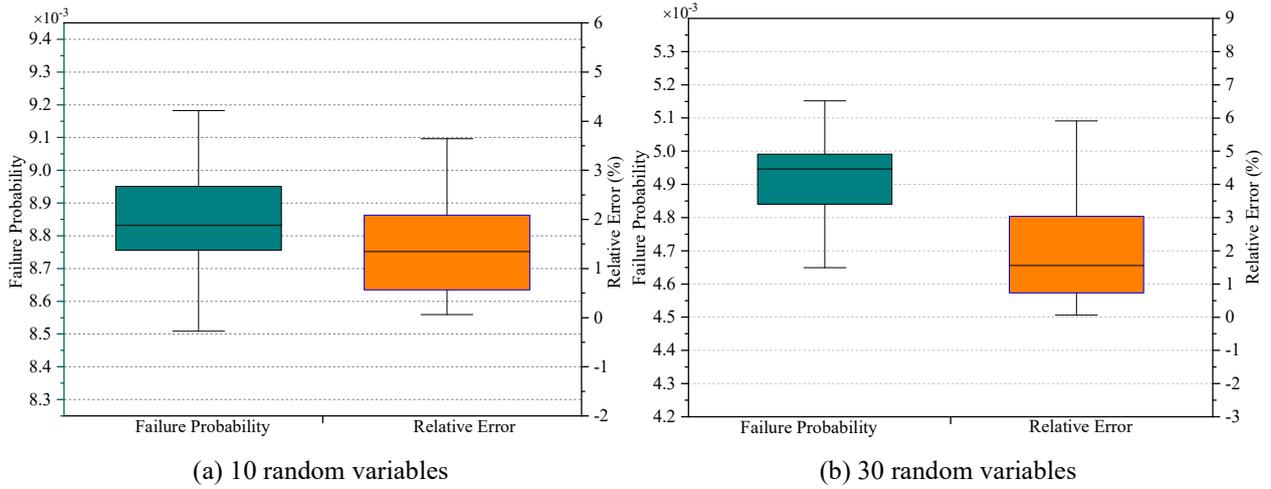

(a) 10 random variables

(b) 30 random variables

**Fig. 13**. Boxplots of predicted failure probability and relative error (CFAK-H-MCS).

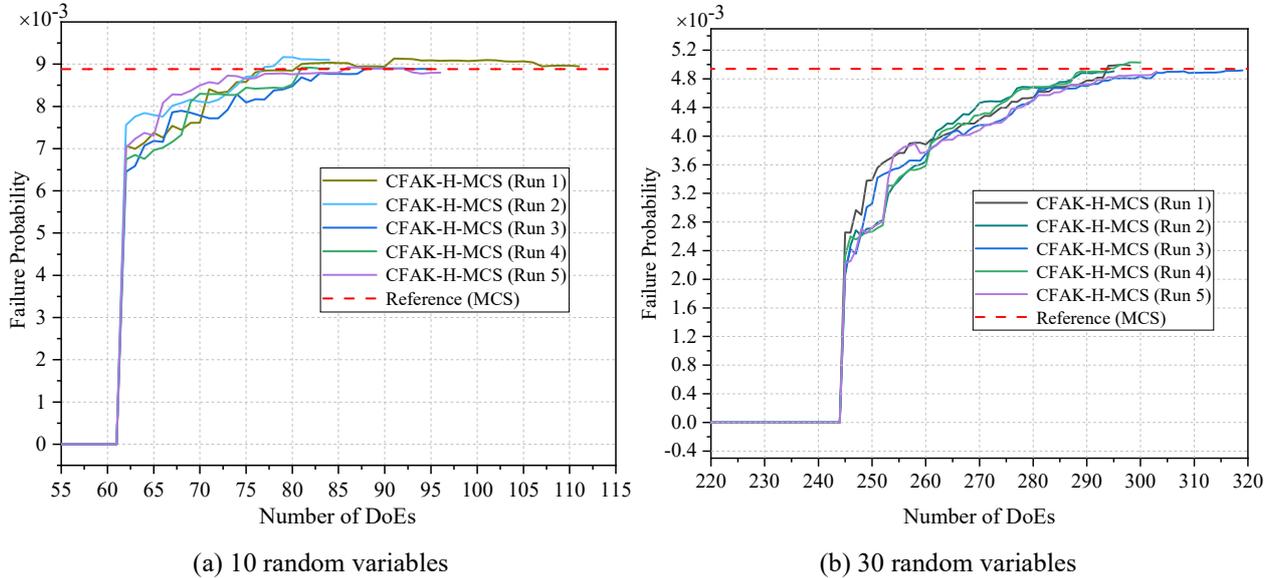

(a) 10 random variables

(b) 30 random variables

**Fig. 14**. Predicted failure probability along with the number of DoE samples (CFAK-H-MCS).

# 5 Conclusions

This study develops an AL surrogate model method based on Kriging-HDMR modeling for reliability analysis. The proposed approach facilitates the approximation of high-dimensional LSFs through a composite representation constructed from multiple low-dimensional sub-surrogate models. The architecture of the surrogate modeling framework comprises three distinct stages: developing single-variable sub-surrogate models for all random variables, identifying the requirements for coupling-variable sub-surrogate models, and constructing the coupling-variable sub-surrogate models. Regarding the selection of DoE samples, corresponding optimization mathematical models are formulated based on the characteristics of each modeling stage, with optimization objectives incorporating the effects of uncertainty variance, predicted mean, sample point location and inter-sample distances. In addition, the CSP-free approach is adopted to achieve the selection of informative samples. The performance of the proposed method is evaluated through a series of numerical experiments. The following key conclusions are drawn:

(1) The combination of HDMR and AL-based Kriging surrogate model provides an effective and efficient framework for high-performance reliability problems. Numerical results demonstrate that the proposed method achieves both high computational efficiency and strong predictive accuracy in the context of reliability analysis.

(2) Within the Kriging-HDMR framework, the progressive modeling strategy, which sequentially constructs sub-surrogate models in ascending order of complexity, not only enhances the logical consistency of the algorithm but also ensures that the higher-order sub-models contribute meaningfully to the overall approximation of the limit state function.

(3) The optimization mathematical model established for informative sample selection is tailored to the characteristics of reliability problems, ensuring the implementation of the AL strategy.

(4) The introduction of PSO-based sample selection strategy for AL significantly overcomes the limitations of conventional CSP-based methods in identifying high informative samples, leading to substantial improvements in the overall performance of the proposed surrogate model method.

**Acknowledgments**

The project is funded by the National Natural Science Foundation of China (Grant No. 52178209, Grant No. 51878299), Guangdong Basic and Applied Basic Research Foundation, China (Grant No. 2025A1515011664, Grant No. 2021A1515012280, Grant No. 2020A1515010611), Science and Technology Innovation Program from Water Resources of Guangdong Province (Grant No. 2025-02) and Innovation Team Project for Ordinary Universities in Guangdong Province (Grant No. 2023KCXTD005).